\documentclass[letterpaper]{article} 
\usepackage[submission]{aaai25}  
\usepackage{times}  
\usepackage{helvet}  
\usepackage{courier}  
\usepackage[hyphens]{url}  
\usepackage{graphicx} 
\usepackage{natbib}  
\usepackage{caption} 
\usepackage{algorithm}
\usepackage{algorithmic}

\usepackage{newfloat}
\usepackage{listings}
\DeclareCaptionStyle{ruled}{labelfont=normalfont,labelsep=colon,strut=off} 
\floatstyle{ruled}
\newfloat{listing}{tb}{lst}{}
\floatname{listing}{Listing}
\usepackage{multirow}
\usepackage{booktabs}
\usepackage{amssymb}
\usepackage{amsmath}
\usepackage{appendix}
\usepackage{xcolor}
\usepackage{bm}
\DeclareMathOperator*{\argmax}{\arg\!\max}
\DeclareMathOperator*{\argmin}{\arg\!\min}
\newcommand{\ourModel}{AnchorInv}
\newcommand{\cut}[1]{}

\usepackage{acronym}
\usepackage{subcaption}
\acrodef{STFT}[STFT]{short-time Fourier transform}
\acrodef{RSSI}[RSSI]{radio signal strength indication}
\acrodef{VOG}[VOG]{video-oculography}
\acrodef{ERP}[ERP]{event-related potential}
\acrodef{EEG}[EEG]{Electroencephalogram}
\acrodef{ECG}[ECG]{Electrocardiogram}
\acrodef{FO}[FO]{face outline}
\acrodef{EDA}[EDA]{electrodermal activity}
\acrodef{BVP}[BVP]{blood volume plse}
\acrodef{PPG}[PPG]{photoplethysmography}
\acrodef{Bio-Z}[Bio-Z]{bio-impedance}
\acrodef{GYR}[GYR]{gyroscope}
\acrodef{EOG}[EOG]{electrooculography}
\acrodef{HR}[HR]{heart rate}
\acrodef{SC}[SC]{skin conductance}
\acrodef{DA}[DA]{data augmentation}
\acrodef{TL}[TL]{transfer learning}
\acrodef{MAML}[MAML]{Model-Agnostic Meta-Learning}
\acrodef{GAN}[GAN]{generative adversarial network}
\acrodef{WGAN}[WGAN]{Wasserstein generative adversarial network}
\acrodef{CGAN}[CGAN]{conditional generative adversarial network}
\acrodef{GRU}[GRU]{gated recurrent unit}
\acrodef{BERT}[BERT]{bidirectional encoder representations from transformers}
\acrodef{MLP}[MLP]{multilayer perceptron}
\acrodef{EHR}[EHR]{electronic health record}
\acrodef{IMU}[IMU]{inertial measurement unit}
\acrodef{SVM}[SVM]{support vector machine}
\acrodef{CNN}[CNN]{convolutional neural network}
\acrodef{LSTM}[LSTM]{long short-term memory}
\acrodef{GNN}[GNN]{graph neural network}
\acrodef{LBP}[LBP]{linear binary pattern}
\acrodef{EMG}[EMG]{electromyography}
\acrodef{BCI}[BCI]{brain computer interface}
\acrodef{SSVEP}[SSVEP]{steady state visually evoked potential}
\acrodef{SoC}[SoC]{system on a chip}
\acrodef{SVR}[SVR]{support vector regression}
\acrodef{GCN}[GCN]{graph convolutional network}
\acrodef{FOMAML}[FOMAML]{first-order model-agnostic meta-learning}
\acrodef{MSE}[MSE]{mean squared error}
\acrodef{LLM}[LLM]{large language model}
\acrodef{NLP}[NLP]{natural language processing}
\acrodef{ACC}[ACC]{accelerometry}
\acrodef{TEMP}[TEMP]{temperature}
\acrodef{FSCIL}[FSCIL]{Few-Shot Class-Incremental Learning}
\acrodef{MAE}[MAE]{Mean Absolute Error}

\title{AnchorInv: Few-Shot Class-Incremental Learning of Physiological Signals via Feature Space-Guided Inversion}
\author{
    Chenqi Li\textsuperscript{\rm 1}\thanks{Correspondence to chenqi.li@eng.ox.ac.uk.}, Boyan Gao\textsuperscript{\rm 1}, Gabriel Davis Jones\textsuperscript{\rm 2}, Timothy Denison\textsuperscript{\rm 1}, Tingting Zhu\textsuperscript{\rm 1}
}
\affiliations{
    \textsuperscript{\rm 1}Department of Engineering Science, University of Oxford, Parks Road, Oxford OX1 3PJ, United Kingdom\\
    \textsuperscript{\rm 2}Oxford Digital Health Labs, Nuffield Department of Women’s \& Reproductive Health, University of Oxford, Women’s Centre, John Radcliffe Hospital, Oxford OX3 9DU, United Kingdom

}

\begin{document}

\maketitle

\begin{abstract}
Deep learning models have demonstrated exceptional performance in a variety of real-world applications. These successes are often attributed to strong base models that can generalize to novel tasks with limited supporting data while keeping prior knowledge intact. However, these impressive results are based on the availability of a large amount of high-quality data, which is often lacking in specialized biomedical applications. In such fields, models are usually developed with limited data that arrive incrementally with novel categories. This requires the model to adapt to new information while preserving existing knowledge. Few-Shot Class-Incremental Learning (FSCIL) methods offer a promising approach to addressing these challenges, but they also depend on strong base models that face the same aforementioned limitations. To overcome these constraints, we propose AnchorInv following the straightforward and efficient buffer-replay strategy. Instead of selecting and storing raw data, AnchorInv generates synthetic samples guided by anchor points in the feature space. This approach protects privacy and regularizes the model for adaptation. When evaluated on three public physiological time series datasets, AnchorInv exhibits efficient knowledge forgetting prevention and improved adaptation to novel classes, surpassing state-of-the-art baselines.

\end{abstract}

\section{Introduction}

The rapid advancement of wearable technologies and the ubiquity of physiological data have sparked great excitement in the development of data-driven models to improve health analytics \cite{ravi2016deep, miotto2018deep}. From clinical diagnostics to brain-computer interfaces and fitness tracking, AI models are revolutionizing how we approach healthcare and interact with the world \cite{zhang2019survey, zhang2022deep, ramanujam2021human}. Although data-driven models have shown incredible performance in a variety of benchmarks, they rely on large amounts of training data and can only perform very specific tasks \cite{sun2017revisiting}. New classes may arise when users demand additional functionalities, or new diseases emerge. Furthermore, access to extensively labeled datasets for emerging classes faces many challenges, such as the long-tail distribution of rare diseases, the high cost of data collection and annotation, as well as privacy and ethical considerations \cite{gao2023addressing}. Therefore, it is paramount for data-driven models to possess the ability to adapt to new classes with limited supporting samples, mimicking how humans leverage past experiences to learn new tasks with minimal guidance. 

\ac{FSCIL} presents a realistic challenge for models to learn new classes with few-shot samples, without forgetting old classes \cite{tao2020few}. Existing \ac{FSCIL} methodologies rely on two important assumptions: a large number of base classes to train a generalizable backbone and the ability to store samples in memory for access in incremental sessions. For biomedical time series datasets, the number of base classes available is limited. A backbone trained with limited classes does not generalize well to new classes, which calls for the need to finetune the backbone in incremental sessions. Naive finetuning leads to catastrophic forgetting \cite{goodfellow2013empirical}, in which the model abruptly forgets previously learned knowledge once it has been trained to learn new information. To help regularize prior knowledge during finetuning, many approaches store samples in memory, violating data privacy concerns. To address these challenges, we investigate inversion-based methods for \ac{FSCIL} application to physiological time series datasets. We propose to guide the inversion process in the feature space, as opposed to the label space, in order to provide a faithful and diverse representation of prior classes. Our contributions are summarized as follows:
\begin{itemize}
    \item We empirically show that as the number of classes available to train the backbone reduces, the generalization capability of the backbone to unseen categories decreases. This motivates the need to finetune the backbone during incremental sessions for physiological datasets, which often have a limited number of base classes.
    \item We propose feature space-guided model inversion to retroactively synthesize diverse and representative samples of previously seen classes. The inverted samples enable the finetuning of the model backbone to incrementally learn new classes while maintaining a faithful representation of previous knowledge and adhering to data privacy and sharing regulations.
    \item We further investigate how the choice of anchor points from base classes affects few-shot class-incremental learning performance.
    \item We point out that the existing evaluation of \ac{FSCIL} methods is highly sensitive to the choice of incremental training samples due to its few-shot nature. To address this, we randomly sample multiple training sets for each incremental session and repeat the adaption process across multiple trials to provide an expectation of model performance in the wild. We comprehensively benchmark our proposed method against SOTA methods on three physiological datasets, surpassing state-of-the-art baselines.
\end{itemize}

\begin{figure*}[t]
    \centering
    \includegraphics[width=\linewidth]{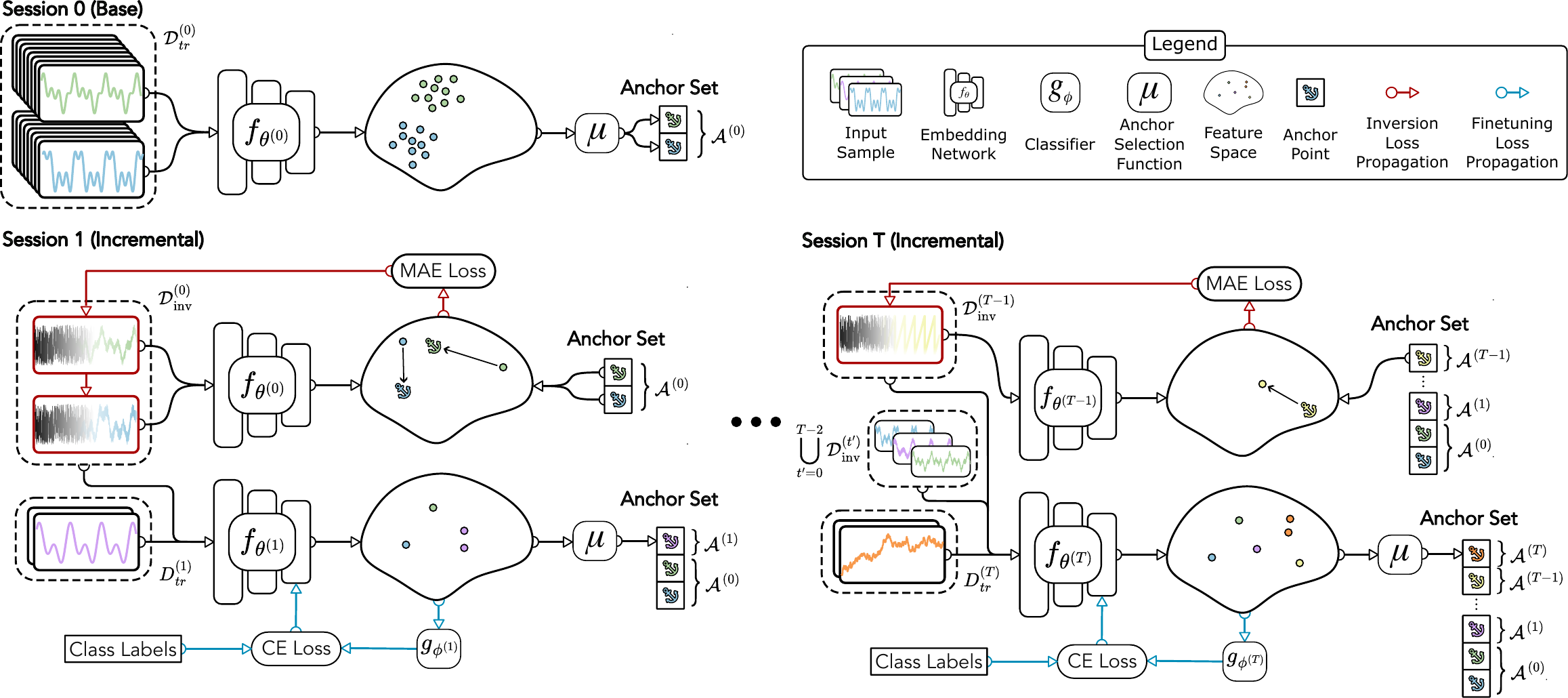}
    \caption{\textbf{Overview of AnchorInv}. In the base session, the training dataset is projected to the feature space, and anchor points for each class are identified and saved in the anchor set memory. In incremental sessions, the anchor set guides the model inversion process to generate representative samples of previously seen classes. The inverted samples and the few-shot training set are subsequently used to finetune the backbone. Anchor points for the new classes from the session are also appended to the anchor set memory, for access in later incremental sessions.}
    \label{fig:anchorinvoverview}
\end{figure*}

\section{Background and Related Works}

\subsection{Problem Definition}
In FSCIL, a learner is deployed to handle a sequence of $T$ learning tasks, $\{\mathcal{D}^{(0)}_{tr}, \mathcal{D}^{(1)}_{tr}, \ldots, \mathcal{D}^{(T)}_{tr}\}$, and required to retain the acquired knowledge without forgetting. In the base session, a sufficient amount of data is available for a given set of base classes as $\mathcal{D}^{(0)}_{tr}$. In each incremental session, the learner addresses a $N$-way $K$-shot problem by observing $N$ sample-label pairs for $K$ incremental classes: $\mathcal{D}^{(t)}_{tr} = \{(\mathbf{x}_{i}^{(t)}, y_{i}^{(t)})\}_{i=1}^{N\times K}$, with corresponding label space $\mathcal{C}^{(t)}$ that satisfies $\mathcal{C}^{(t)} \cap \mathcal{C}^{(t')} = \varnothing$, $\forall t \neq t'$. Note that for session $t$, all prior and future datasets are inaccessible, with $\mathcal{D}^{(t)}_{tr}$ being the only accessible dataset for session $t$. After adaptation, the model is evaluated on an unseen test set that consists of samples from all seen classes to date.

\subsection{Related Works}

\subsubsection{Few-Shot Class Incremental Learning}
Recently, Sun et al. pioneered the exploration of \ac{FSCIL} for physiological signal \cite{sun2023few}. The authors proposed a MAPIC framework, which consists of a meta-learning trained feature extractor, an attention-based prototype enhancement module, and a composition of three loss functions to adapt incremental classes \cite{sun2023few}. Concurrently, Ma et al. proposed a graph convolutional network with a growing linear classifier for \ac{EEG} emotion recognition \cite{ma2023fewa}. While \ac{FSCIL} for physiological signals has been limited to these two works, \ac{FSCIL} for images proposed a broader range of approaches.

A popular choice for \ac{FSCIL} in the computer vision domain focuses on training a robust backbone that can generalize well to new classes. Forward-compatible methods seek to encourage intra-class compactness and reserve space for incremental classes during base-stage training, by creating virtual classes \cite{peng2022few, song2023learning} or simulating pseudo-incremental sessions \cite{zhou2022few, zhou2022forward, chi2022metafscil}. Many methods also assume that a robust and generalizable backbone is available from base-stage training. During incremental sessions, the backbone is frozen and performance gains are achieved through the design of classifiers \cite{yang2023neural, zhuang2023gkeal, hersche2022constrained} or calibrating prototypes \cite{yao2022few, wang2024few}. However, in a healthcare setting, the number of base classes, often in single digits, is much less than the 60-100 base classes available in computer vision datasets. The lack of sufficient base classes cannot support the creation of pseudo-incremental sessions or the synthesis of new classes through interpolation between existing ones. Furthermore, identifying relevant augmentations to synthesize pseudo-classes that effectively simulate real incremental classes for different physiological modalities is rather non-trivial. Moreover, a backbone trained on limited base-stage data does not generalize well to new classes, limiting the effectiveness of methods that freeze the backbone. 

Many works also focused on finetuning the backbone to improve incremental learning performance while maintaining base knowledge. Notably, some have focused on identifying unimportant weights of the backbone network to finetune \cite{kang2023soft, kim2023warping}, but their performances have been shown to be inferior to the aforementioned methods. To provide better regularization for base class knowledge, many methods store exemplars from the base session and finetune using a combination of stored and new class data \cite{zhao2023few, chen2021incremental, shi2021overcoming, dong2021few, kukleva2021generalized, akyurek2021subspace, zhu2022feature}. Unfortunately, due to strict data regulations in a healthcare setting, direct storage of samples raises data and privacy concerns. To address this challenge, generative replay methods mitigate privacy concerns by training an auxiliary generator to synthesize representative samples from prior classes \cite{agarwal2022semantics, shankarampeta2021few, liu2022few}. Training of \acp{GAN} incur significant complexity \& computation overhead and encounter many challenges such as mode collapse and vanishing gradients \cite{arjovsky2017towards}. 

An alternative to \ac{GAN} is model inversion techniques such as DeepDream \cite{mordvintsev2015inceptionism} and DeepInv \cite{yin2020dreaming}. These generative models directly extract training-like samples from pre-trained deep models without the need for an additional generator and discriminator, resulting in a more efficient and stable sample generation process. When applied to \ac{FSCIL}, this property helps guide regularization in the label space by minimizing classification loss. However, this indirect regularization in the feature space may not sufficiently prevent catastrophic forgetting, especially given the non-parametric classifier design commonly used in \ac{FSCIL}. To address this, we generate replay samples through embedding anchors that characterize the feature space, thereby preventing significant shifts in the feature space when the model is optimized to integrate new tasks.

\begin{table}[!h] 
\centering
\begin{tabular}{ccc} 
\toprule 
\textbf{\begin{tabular}[c]{@{}l@{}}Number of \\ Base Classes\end{tabular}} & \textbf{ProtoNet} & \textbf{\begin{tabular}[c]{@{}l@{}}Random \\ Chance\end{tabular}} \\
\midrule
10                                                                         &                   41.62 $\pm$ 2.49 & 6.25 $\pm$ 0.00                                                             \\
8                                                                          &                   39.53 $\pm$ 2.27 & 7.14 $\pm$ 0.00                                                             \\
6                                                                          &                   40.16 $\pm$ 1.90 & 8.33 $\pm$ 0.00                                                             \\
4                                                                          &                   32.27 $\pm$ 1.72 & 10.00 $\pm$ 0.00                                                            \\
\bottomrule

\end{tabular}
\caption{\textbf{Impact of Reduced Base Classes on GRABMyo Dataset}: We train ProtoNet with varying numbers of base classes and compare the macro-F1 score (\%) over 6 unseen classes. Random chance denotes a theoretical baseline that randomly guesses all predictions.}
\label{tab:reducebase}
\end{table}

\subsection{Impact of Reduced Base Classes}

To show how the number of base classes available impacts the performance of the generalization of the backbone, we train a prototypical network with \{10, 8, 6, 4\} base classes, and directly transfer it for inference on six unseen classes. We repeat the evaluation 20 times with different incremental training sets randomly sampled and report the mean and standard deviation. As seen in Table \ref{tab:reducebase}, as the number of base classes decreases, the performance on new classes decreases, despite the problem becoming easier with fewer classes to choose from, as reflected in the increase in random chance performance. This reinforces the importance of finetuning the backbone in incremental sessions, when there are limited base classes to train the backbone.

\section{Method}
An overview of our method is summarized in Figure \ref{fig:anchorinvoverview} and Algorithm \ref{alg:algorithm}.

\begin{algorithm}[]
\caption{\textbf{Pseudocode for AnchorInv}}
\label{alg:algorithm}
\textbf{Input}: $\{\mathcal{D}^{(t)}_{tr}\}_{t=0}^{T}$ \\
\textbf{Output}: $\bm{\theta}^{(T)}, \bm{\phi}^{(T)}$ \\
\textbf{Base Session}
\begin{algorithmic}[1] 
\STATE Randomly initialize $\bm{\theta}^{(0)}, \bm{\phi}^{(0)}$
\STATE Train $\bm{\theta}^{(0)}, \bm{\phi}^{(0)}$ on $\mathcal{D}^{(0)}_{tr}$
\STATE Replace $\bm{\phi}^{(0)}$ with prototype vectors using Eq. \ref{eq:proto}
\STATE Calculate $\mathcal{A}^{(0)}$ using $\mathcal{D}^{(0)}_{tr}$, following Eq. \ref{eq:proj1}-\ref{eq:anchor2}
\end{algorithmic}

\textbf{Incremental Sessions}
\begin{algorithmic}[1] 
\FOR{session $(t+1) \gets 1$ to $T$}
\STATE \color{gray} [Optional] Initialize $\bm{\phi}^{(t)}$ for incremental classes as class prototype from $\mathcal{D}^{(t+1)}_{tr}$ following Eq. \ref{eq:proto} \color{black}
\STATE Invert $\mathcal{D}_{\text{inv}}^{(t)}$ using $\mathcal{A}^{(t)}$ following Eq. \ref{eq:inv2}
\STATE Calculate $\mathcal{L}_{\text{old}}$ using $\cup_{t'=0}^{t}  \mathcal{D}_{\text{inv}}^{(t')}$ following Eq. \ref{eq:lossold} 
\STATE Calculate $\mathcal{L}_{\text{new}}$ using $\mathcal{D}^{(t+1)}_{tr}$ following Eq. \ref{eq:lossnew} 
\STATE Calculate $\mathcal{L}_{\text{ft}}$ following Eq. \ref{eq:lossft} 
\STATE $\bm{\theta}^{(t+1)} \leftarrow \bm{\theta}^{(t)} - \eta \nabla_{\bm{\theta}^{(t)}}  \mathcal{L}_{\text{ft}} $
\STATE $\bm{\phi}^{(t+1)} \leftarrow \bm{\phi}^{(t)} - \eta \nabla_{\bm{\phi}^{(t)}}  \mathcal{L}_{\text{ft}} $
\STATE Calculate $\mathcal{A}^{(t+1)}$ using $\mathcal{D}^{(t+1)}_{tr}$ following Eq. \ref{eq:proj1}-\ref{eq:anchor2}
\ENDFOR

\end{algorithmic}
\end{algorithm}

\subsection{Model Inference} 
\label{sec:model_inference}

Define the inference process for a neural network as:
\begin{align}
\mathbf{h}_{i}^{(t)} &= f_{\bm{\theta}^{(t)}} \left( \mathbf{x}_{i}^{(t)} \right) \\
\hat{y}_{i}^{(t)} &= g_{\bm{\phi}^{(t)}} \left( \mathbf{h}_{i}^{(t)} \right)
\end{align}
where $\mathbf{x}_{i}^{(t)} \in \mathbb{R}^{H \times W}$ denotes the $i$-th training sample from session $t$ with $H$ channels and $W$ time steps. $f_{\bm{\theta}^{(t)}}$ and $g_{\bm{\phi}^{(t)}}$ denote the backbone network and classifier parametrized by $\bm{\bm{\theta}}$ and $\bm{\phi}$ from the $t$-th session. $\mathbf{h}_{i}^{(t)} \in \mathbb{R}^{D}$ represents the feature or embedding vector with dimension $D$. $\hat{y}_{i}^{(t)}$ denotes the model prediction for the corresponding training sample. 
We use a metric-based classifier:
\begin{align} 
    g_{\bm{\phi}^{(t)}} \left( \mathbf{h}_{i}^{(t)} \right) &:= \argmax_k s \left( \mathbf{h}_{i}^{(t)}, \bm{\phi}^{(t)}_k \right) \label{eq:protopred1} \\
    s \left( \mathbf{h}_{i}^{(t)}, \bm{\phi}^{(t)}_k \right) &= \frac{\text{exp} \left(-d(\mathbf{h}_{i}^{(t)}, \bm{\phi}^{(t)}_k) / T \right)}{\sum\limits_{k'=1}^{\cup^{t}_{c=0}{\mathcal{C}^{(c)}}}\text{exp} \left( -d(\mathbf{h}_{i}^{(t)}, \bm{\phi}^{(t)}_{k'}) / T \right) } \label{eq:protopred2} \\
    d(\mathbf{h}_{i}^{(t)}, \bm{\phi}^{(t)}_k) &= - \frac{\mathbf{h}_{i}^{(t)} \cdot \bm{\phi}^{(t)}_k}{\|\mathbf{h}_{i}^{(t)}\| \|\bm{\phi}^{(t)}_k\|} \label{eq:protopred3}
\end{align}

where $T$ is a temperature parameter. $d(\cdot, \cdot)$ denotes the distance function that measures the similarity between the pair of embeddings, and in our case, the negative cosine distance is applied with a detailed description in Equation \ref{eq:protopred3} where $\| \cdot \|$ denotes $l$2-norm. $\bm{\phi}^{(t)}_{k}$ denotes classifier weight for class $k$ in session $t$ and is trained jointly with $\bm{\theta}^{(t)}$. Once the backbone network has been trained, the classifier weights are replaced as the prototype vector or the average embedding of each class:
\begin{equation} \label{eq:proto}
    \bm{\phi}^{(t)}_k = \frac{1}{|\mathcal{D}^{(t)}_{tr, k}|} \sum_{(\mathbf{x}_{i}^{(t)}, y_{i}^{(t)}) \in \mathcal{D}^{(t)}_{tr, k} }{f_{\bm{\theta}^{(t)}} \left( \mathbf{x}_{i}^{(t)} \right)}
\end{equation}
where $\mathcal{D}^{(t)}_{tr, k}$ denotes the subset of $\mathcal{D}^{(t)}_{tr}$ that belongs to class $k$, $|\mathcal{D}^{(t)}_{tr, k}|$ denotes the number of samples of the subset, and $|\mathcal{D}^{(t)}_{tr, k}| = N$ for $N$-way-$K$-shot setup. 

\subsection{Feature Space-Guided Inversion}

\subsubsection{Anchor Point Selection}
To enable fine-grained control over the inversion process, we now introduce feature space-guided inversion. Given the training dataset for session $t$, $\mathcal{D}_{tr}^{(t)}$, we can project each input sample into the feature space using the feature extractor $f_{\bm{\theta}^{(t)}}$ and obtain the corresponding feature set:
\begin{align}
    \mathcal{H}^{(t)}_{tr} &= \left\{ \left( \mathbf{h}_{i}^{(t)}, y_{i}^{t}   \right) \right\}_{i=1}^{N}, \label{eq:proj1} \\
    \mathbf{h}_{i}^{(t)} &= f_{\bm{\theta}^{(t)}} \left( \mathbf{x}_{i}^{(t)} \right), \,\,\,  \left( \mathbf{x}_{i}^{(t)}, y_{i}^{(t)} \right) \in \mathcal{D}^{(t)}_{tr} \label{eq:proj2}
\end{align} 
A set of anchor points for session $t$ can be subsequently calculated to summarize the feature set:
\begin{equation} \label{eq:anchor1}
    \mathcal{A}^{(t)} = \mu \left( \mathcal{H}^{(t)}_{tr} \right)
\end{equation}
\begin{equation} \label{eq:anchor2}
    \mathcal{A}^{(t)} = \left\{ \left( \mathbf{a}_{j}^{(t)}, y_{j}^{t} \right)   \right\}_{j=1}^J
\end{equation}

where $\mu$ is an algorithm that calculates a set of representative samples $\mathcal{A}^{(t)}$ from a set of feature vectors $\mathcal{H}^{(t)}_{tr}$. Examples of $\mu$ include random sampling or clustering to identify cluster centroids. $J$ is the size of the anchor set, and $J=P\times K$, if we select $P$ anchor points for each of the $K$ incremental classes.  Intuitively, the anchor set summarizes the distribution of the feature set. 
\subsubsection{Inversion}
We derive a replay set, $\mathcal{D}_{\text{inv}}^{(t)} = \{ ( \hat{\mathbf{x}}_j^{(t)}, y_j^{(t)})\}$, from the selected anchor points of session $t$ to regularize the adaptation of the feature extractor to session $t+1$. A natural formalization of this derivation process is to align the embedding of a potential replay sample, $\mathbf{x}_j^{(t)}$, to an anchor point following the minimization operator:
\begin{equation} \label{eq:inv2}
    \hat{\mathbf{x}}_j^{(t)} = \argmin_{\hat{\mathbf{x}}_j} \mathcal{L} \left( f_{\bm{\theta}^{(t)}} \left( \hat{\mathbf{x}}_j^{(t)} \right), \mathbf{a}_j^{(t)} \right)    
\end{equation}
where $\mathcal{L}$ represents \ac{MAE}. Optimization can be performed via gradient descent on randomly initialized replay samples.

\subsubsection{Model Finetuning}
The objectives for finetuning the model are:
\begin{align} 
    \mathcal{L}_{\text{ft}} &= \mathcal{L}_{\text{new}} + \lambda \mathcal{L}_{\text{old}}  \label{eq:lossft} \\
    \mathcal{L}_{\text{old}} &= \mathcal{L} \left( g_{\bm{\phi}^{(t)}} \circ  f_{\bm{\theta}^{(t)}} \left( \hat{\mathbf{x}}_j^{(t)} \right)  ,y_j^{(t)} \right)\label{eq:lossold} \\
    \mathcal{L}_{\text{new}} &= \mathcal{L} \left( g_{\bm{\phi}^{(t)}} \circ f_{\bm{\theta}^{(t)}} \left( \mathbf{x}_i^{(t)} \right),y_i^{(t)} \right) \label{eq:lossnew} 
\end{align}
where $\mathcal{L}_{\text{old}}$ regularizes knowledge of all previously seen classes using inverted samples from $\left( \hat{\mathbf{x}}_j, y_j \right) \in  \cup_{t'=0}^{t}  \mathcal{D}_{\text{inv}}^{(t')}$ and $\mathcal{L}_{\text{new}}$ learns the new classes using the training dataset for session $t$ from $\left( \mathbf{x}_{i}^{(t)}, y_{i}^{(t)} \right) \in \mathcal{D}^{(t)}_{tr}$. The model update then follows:

\begin{equation*}
    \bm{\theta}^{(t+1)} \leftarrow \bm{\theta}^{(t)} - \eta \nabla_{\bm{\theta}^{(t)}}   \mathcal{L}_{\text{ft}} 
\end{equation*}
\begin{equation*}
    \bm{\phi}^{(t+1)} \leftarrow \bm{\phi}^{(t)} - \eta \nabla_{\bm{\phi}^{(t)}}   \mathcal{L}_{\text{ft}}
\end{equation*}

In summary, feature space-guided inversion summarizes the distribution of each session in the feature space as an anchor set. The anchor set then serves as the target for inversion to synthesize representative samples for all previously learned classes. This enables the finetuning of the model with incremental class samples while maintaining knowledge over prior classes and safeguarding data privacy.

\subsection{Evaluation across trials}

\begin{figure}[!h]
    \centering
    \includegraphics[width=0.8\linewidth]{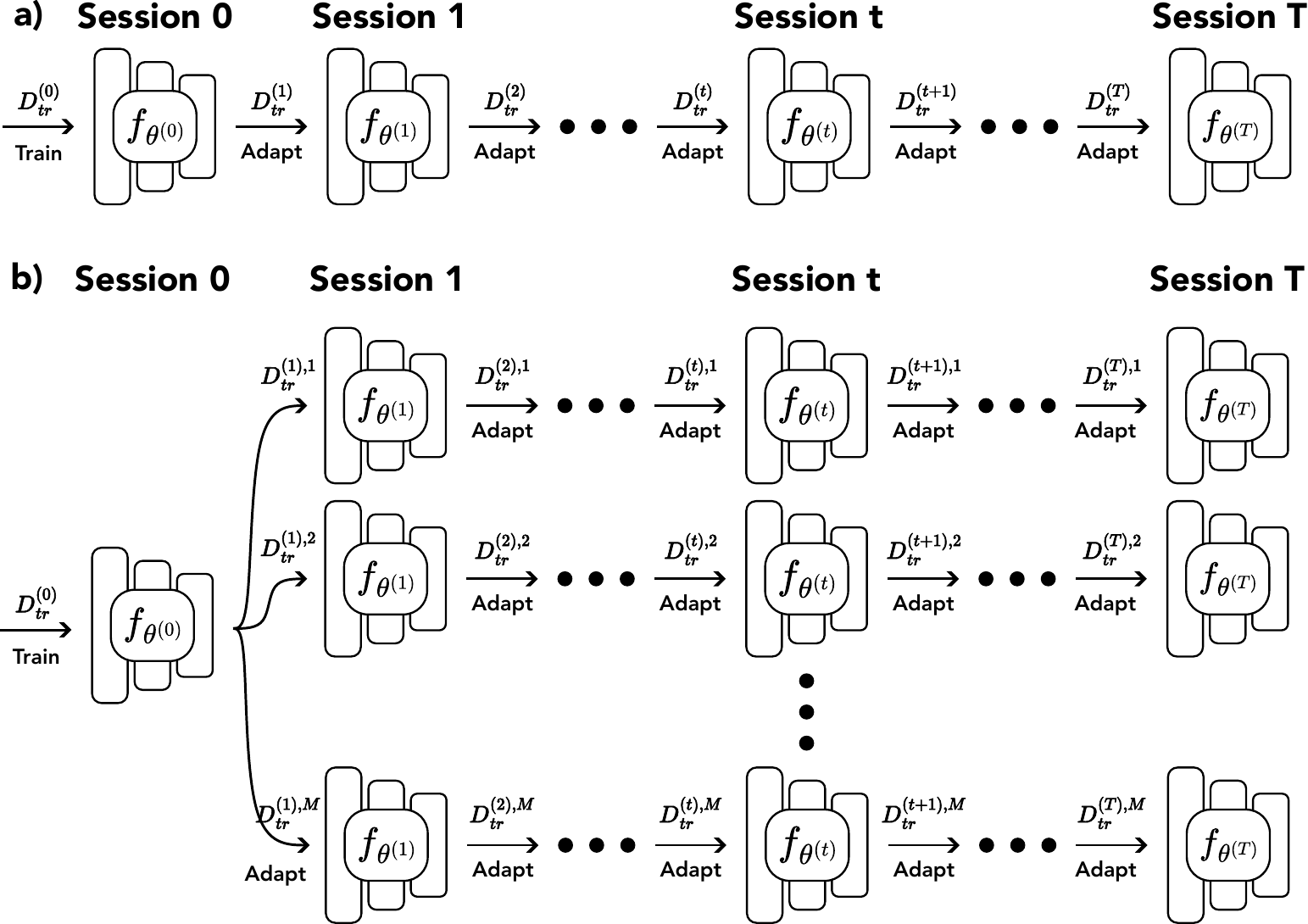}
    \caption{\textbf{Overview of \ac{FSCIL} Evaluation Procedure} a) illustrates the FSCIL evaluation process in the existing literature. b) illustrates the FSCIL evaluation process adopted in this work. In Session 1, we create $M$ copies of the base session model. In each incremental session, each copy is adapted with a randomly sampled training set. Each copy is evaluated on the test set and the mean and standard deviation are reported across all $M$ trials.}
    \label{fig:trial_eval}
\end{figure}

Given the few-shot nature of incremental training sets, the selection of few-shot samples can yield significant variation in performance across different methods. This is further exacerbated for physiological datasets, where inter-subject variability can lead to large variances in samples. It is commonly true that during deployment, practitioners often have to work with the few-shot samples at hand and do not have the opportunity to select which few-shot samples to use. However, in \ac{FSCIL} benchmarks, the incremental training sets are artificially sampled from a larger pool of samples. To remove stochasticity in performance due to the random selection of few-shot samples and to provide an expectation of the model's performance in the wild, we sample $M$ incremental training sets in each session. The $M$ adapted models are then evaluated in the test set and we report the mean and standard deviation for all $M$ trials. This process is visually illustrated in Figure \ref{fig:trial_eval}.

\section{Experiments} 
We evaluate \ourModel{} on several \ac{FSCIL} scenarios simulated from three standard physiological datasets with both \ac{EEG} and \ac{EMG} data. For a comprehensive assessment, the classification tasks are further diversified with varying categories, numbers of base classes, incremental classes, and different levels of class correlation. A detailed description of train-test split and preprocessing can be found in Appendix \ref{sec:preprocessing}.
\subsection{Dataset and Setup}
\subsubsection{Brain-Computer Interface Motor-Imagery}
The BCI-IV 2a dataset \cite{brunner2008bci} is an \ac{EEG} dataset for the classification of 4 distinct motor imagery classes. We select the first two classes as base classes (left hand, right hand), and the two incremental classes (feet, tongue) are learned over two incremental sessions, with a 1-way-10-shot setup.
\subsubsection{Grading of Neonatal Hypoxic-Ischemic Encephalopathy}
The NHIE dataset \cite{o2023neonatal} is an \ac{EEG} dataset for grading the severity of background abnormalities in neonates who have been diagnosed with hypoxic-ischaemic encephalopathy, a common brain injury as a result of impaired oxygen or blood flow to the brain during the time of birth. The severity is categorized into 4 grades, ranging from normal to inactive, following the guidelines from \cite{murray2009early}. Unlike the other datasets, adjacent grades may share similar characteristics. We select the first 2 classes as base classes (normal/mildly abnormal, moderate abnormalities), and the remaining 2 incremental classes (major abnormalities, inactive) are learned over 2 incremental sessions, with a 1-way-10-shot setup.
\subsubsection{Gesture Recognition}
The GRABMyo dataset \cite{jiang2022gesture} is an \ac{EMG} dataset for classifying 16 distinct hand and finger gestures. We select the first 10 classes as base classes, and learn 6 incremental classes over 6 incremental sessions, with a 1-way-10 shot setup.

\subsection{Implementation Details}
\subsubsection{Network Architecture}
For the embedding network, $f_{\bm{\theta}^{(t)}}$, we use the convolution-transformer backbone of EEG-Conformer \cite{song2023eeg}, a state-of-the-art architecture for \ac{EEG} classification. The backbone consists of a convolution module that captures one-dimensional temporal and spatial information, followed by a transformer module to extract global correlation. For the classifier, $g_{\bm{\phi}^{(t)}}$, we follow the classifier described in Equations \ref{eq:protopred1} to \ref{eq:protopred3} with $T=16$. For the BCI dataset, we adopt the original EEG-conformer architecture without any modifications. For the NHIE and GRABMyo datasets, given different input sizes, we modify the convolution module so that the feature vector, $\mathbf{h}_{i}^{(t)} \in \mathbb{R}^{D}$, are similar in dimension and the transformer module is adopted without any modification. Detailed convolution architecture is summarized in Appendix \ref{sec:conv_module_architecture}, resulting in feature vectors dimensions $D=2440$ for BCI, $D=2360$ for NHIE,  $D=2320$ for GRABMyo. 

\subsubsection{Base Session Training and Incremental Session Adaptation}
Please refer to Appendix \ref{sec:trainadapt} for our implementation details on model training and adaption.

\subsection{Benchmark Comparison}

\begin{table*}[!h] 
\centering

\begin{subtable}[t]{0.95\textwidth}
\subcaption{\textbf{BCI 1-way-10-shot \ac{FSCIL} across 100 trials (\%)}}
\label{tab:bci_results}
\resizebox{1\textwidth}{!}{
\begin{tabular}{cccccccccc}
\toprule
\toprule
\multicolumn{1}{c}{\multirow{2}{*}{\textbf{Method}}} & \multicolumn{3}{c}{\textbf{Macro-F1 (All Classes)}}      & \multicolumn{2}{c}{\textbf{Macro-F1 (Base Classes)}} & \multicolumn{2}{c}{\textbf{Macro-F1 (Incremental Classes)}} \\
\cmidrule(lr){2-4} \cmidrule(lr){5-6}\cmidrule(lr){7-8}
\multicolumn{1}{c}{}                           & \textbf{Base Session} & \textbf{Session 1} & \textbf{Session 2} & \textbf{Session 1}   & \textbf{Session 2}   & \textbf{Session 1}    & \textbf{Session 2}    \\
\midrule
 Finetune &        78.81 & 42.39 $\pm$ 5.40 & 27.83 $\pm$ 3.50 & 49.40 $\pm$ 11.56 &  41.89 $\pm$ 9.78 & 28.36 $\pm$ 8.65 & 13.77 $\pm$ 3.68 \\
    NC-FSCIL &        78.39 & 48.77 $\pm$ 1.80 & 34.85 $\pm$ 1.53 &  60.32 $\pm$ 1.56 &  51.04 $\pm$ 1.44 & 25.67 $\pm$ 5.71 & 18.66 $\pm$ 3.08 \\
 ProtoNet &        77.17 & 45.36 $\pm$ 0.80 & 31.02 $\pm$ 0.94 &  \textbf{61.52 $\pm$ 0.35} &  \textbf{51.76 $\pm$ 0.32} & 13.03 $\pm$ 2.35 & 10.28 $\pm$ 1.76 \\
     TEEN &        77.17 & 46.48 $\pm$ 1.08 & 32.69 $\pm$ 1.24 &  51.05 $\pm$ 0.74 &  51.38 $\pm$ 0.66 & 17.34 $\pm$ 3.58 & 13.99 $\pm$ 2.52 \\
DeepDream &        77.17 & 49.97 $\pm$ 2.03 & 32.34 $\pm$ 1.29 &  58.08 $\pm$ 3.46 &  50.16 $\pm$ 2.66 & 33.77 $\pm$ 5.97 & 14.52 $\pm$ 2.10 \\
  DeepInv &        77.17 & 50.16 $\pm$ 1.80 & 32.44 $\pm$ 1.14 &  58.61 $\pm$ 2.94 &  50.51 $\pm$ 2.20 & 33.25 $\pm$ 5.83 & 14.36 $\pm$ 2.06 \\
AnchorInv &        77.17 & \textbf{50.70 $\pm$ 1.45} & \textbf{37.04 $\pm$ 1.24} &  57.52 $\pm$ 2.02 &  48.45 $\pm$ 2.07 & \textbf{37.05 $\pm$ 4.50} & \textbf{25.63 $\pm$ 2.46} \\
\bottomrule
\bottomrule
\end{tabular}%
}
\end{subtable}

\begin{subtable}[t]{0.95\textwidth}
\subcaption{\textbf{NHIE 1-way-10-shot \ac{FSCIL} across 100 trials (\%)}}
\label{tab:nhie_results}
\resizebox{1\textwidth}{!}{
\begin{tabular}{cccccccccc}
\toprule
\toprule
\multicolumn{1}{c}{\multirow{2}{*}{\textbf{Method}}} & \multicolumn{3}{c}{\textbf{Macro-F1 (All Classes)}}      & \multicolumn{2}{c}{\textbf{Macro-F1 (Base Classes)}} & \multicolumn{2}{c}{\textbf{Macro-F1 (Incremental Classes)}} \\
\cmidrule(lr){2-4} \cmidrule(lr){5-6}\cmidrule(lr){7-8}
\multicolumn{1}{c}{}                           & \textbf{Base Session} & \textbf{Session 1} & \textbf{Session 2} & \textbf{Session 1}   & \textbf{Session 2}   & \textbf{Session 1}    & \textbf{Session 2}    \\
\midrule

     Finetune &        82.16 & 24.45 $\pm$ 8.49 &  6.83 $\pm$ 1.17 & 22.14 $\pm$ 10.72 &   0.40 $\pm$ 1.15 &  29.07 $\pm$ 4.46 &  13.21 $\pm$ 1.82 \\
        NC-FSCIL &        81.74 & 53.27 $\pm$ 2.88 & 40.78 $\pm$ 4.84 &  64.17 $\pm$ 6.17 &  57.22 $\pm$ 5.60 &  31.45 $\pm$ 6.88 &  24.35 $\pm$ 5.58 \\
     ProtoNet &        81.83 & 56.07 $\pm$ 2.93 & 60.89 $\pm$ 7.95 &  \textbf{69.05 $\pm$ 3.60} &  68.14 $\pm$ 3.88 &  30.12 $\pm$ 9.16 & 53.64 $\pm$ 16.28 \\
         TEEN &        81.83 &  56.85 $\pm$ 3.0 & 59.76 $\pm$ 7.33 &  68.22 $\pm$ 3.78 &  66.01 $\pm$ 4.59 &  24.10 $\pm$ 8.88 & 53.51 $\pm$ 15.51 \\
    DeepDream &        81.83 & 55.90 $\pm$ 3.73 & 62.72 $\pm$ 4.55 &  67.59 $\pm$ 5.03 &  62.80 $\pm$ 6.76 & 32.51 $\pm$ 12.59 &  62.63 $\pm$ 6.88 \\
      DeepInv &        81.83 & 53.89 $\pm$ 5.36 & 62.79 $\pm$ 2.89 &  68.24 $\pm$ 5.40 &   67.0 $\pm$ 3.63 &  25.20 $\pm$ 9.06 &  58.58 $\pm$ 4.26 \\
    AnchorInv &        81.83 & \textbf{57.01 $\pm$ 2.83} & \textbf{66.03 $\pm$ 2.73} &  68.13 $\pm$ 3.55 &  \textbf{68.67 $\pm$ 2.72} &  \textbf{34.77 $\pm$ 8.79} &  \textbf{63.39 $\pm$ 4.38} \\
\bottomrule
\bottomrule
\end{tabular}%
}
\end{subtable}

\begin{subtable}[t]{0.95\textwidth}
\subcaption{\textbf{GRABMyo 1-way-10-shot \ac{FSCIL} across 20 trials (\%)}}
\label{tab:grabm_results}
\resizebox{1\textwidth}{!}{
\begin{tabular}{ccccccccc}
\toprule
\toprule
\multicolumn{1}{c}{\multirow{2}{*}{\textbf{Method}}} & \multicolumn{7}{c}{\textbf{Macro-F1 (All Classes)}} \\
\cmidrule(lr){2-8}
\multicolumn{1}{c}{}                           & \textbf{Base Session} & \textbf{Session 1} & \textbf{Session 2} & \textbf{Session 3}   & \textbf{Session 4}   & \textbf{Session 5} & \textbf{Session 6}   \\
\midrule

 Finetune &        87.32 & 22.66 $\pm$ 4.32 & 20.23 $\pm$ 3.77 & 18.21 $\pm$ 3.34 &  16.57 $\pm$ 3.00 & 14.87 $\pm$ 2.69 & 13.30 $\pm$ 2.43 \\
    NC-FSCIL &        86.83 &  82.0 $\pm$ 0.99 & 74.07 $\pm$ 1.64 & 69.63 $\pm$ 1.46 & 64.04 $\pm$ 1.61 & 59.96 $\pm$ 1.53 & 58.35 $\pm$ 0.96 \\
 ProtoNet &        86.73 & 82.99 $\pm$ 0.39 & 74.81 $\pm$ 0.69 & 70.77 $\pm$ 0.49 & 65.66 $\pm$ 0.56 & 63.73 $\pm$ 0.60 & 60.15 $\pm$ 0.75 \\
     TEEN &        86.73 & 81.19 $\pm$ 0.56 & 73.42 $\pm$ 0.99 & 69.45 $\pm$ 0.65 & 64.63 $\pm$ 0.81 & 62.33 $\pm$ 7.71 & 59.01 $\pm$ 0.81 \\
DeepDream &        86.73 & 82.84 $\pm$ 1.04 & 74.02 $\pm$ 1.29 & 69.89 $\pm$ 2.11 &  63.71 $\pm$ 2.0 & 58.72 $\pm$ 1.73 & 55.68 $\pm$ 1.74 \\
  DeepInv &        86.73 & 82.83 $\pm$ 1.08 & 73.80 $\pm$ 1.01 & 69.56 $\pm$ 0.80 & 63.48 $\pm$ 1.93 & 58.52 $\pm$ 1.72 & 54.79 $\pm$ 1.30 \\
AnchorInv &        86.73 & \textbf{83.20 $\pm$ 0.70} & \textbf{75.39 $\pm$ 1.01} & \textbf{71.23 $\pm$ 0.79} & \textbf{66.14 $\pm$ 0.65} & \textbf{64.28 $\pm$ 0.64} & \textbf{60.78 $\pm$ 0.66} \\
\bottomrule
\bottomrule
\end{tabular}%
}
\end{subtable}

\caption{\textbf{Comparison of AnchorInv against Baseline}. Across all incremental sessions, AnchorInv is statistically significantly different from all baseline methods with $p < 0.05$ using a two-sided Wilcoxon signed-rank test.}
\label{tab:combined_results}
\end{table*}

\subsubsection{Baseline Methods} 
We conduct a comprehensive comparison with the baseline methods from multiple perspectives. DeepDream \cite{mordvintsev2015inceptionism} and DeepInv \cite{yin2020dreaming}, which share similar motivations with our approach, are implemented and adjusted to substitute our inversion method. Furthermore, we compare with recent state-of-the-art \ac{FSCIL} models, NC-FSCIL \cite{yang2023neural} and TEEN \cite{wang2023few}, both of which have shown outstanding performance in vision tasks. To ensure a fair comparison, we use a backbone with an identical architecture to ours, given the application to a new modality. A non-parametric model, ProtoNet \cite{snell2017prototypical}, which relies heavily on a strong feature extractor, is also included due to its stability across different dataset settings. Furthermore, we denote the model that is directly finetuned on the few-shot samples as `Finetune' in the result tables.

\subsubsection{Evaluation Metric} 
We report the Macro-F1 (All Classes) for all datasets. We also report Macro-F1 (Base Classes) and Macro-F1 (Incremental Classes) for the BCI and NHIE datasets. Macro-F1 (All Classes) presents the average F1 score across all classes seen up to the session. Macro-F1 (Base Classes) measures how well the knowledge of the base class is retained and is defined as the average F1 score across all classes from the base session. Macro-F1 (Incremental Classes) measures how well new classes are learned and is defined as the average F1 score across all incremental classes seen up to the session.
\subsubsection{Performance Comparison}
Experiment results on BCI, NHIE, and GRABMyo are shown in Table \ref{tab:bci_results}, \ref{tab:nhie_results} and \ref{tab:grabm_results} respectively. We note that AnchorInv achieves the best Macro-F1 score in all three datasets throughout all sessions. The improvements are particularly significant in the BCI and NHIE datasets, achieving an improvement of 2.19\% and 3.24\% over the best baseline method. In GRABMyo, the backbone can generalize better to new classes given more base classes, so the room for improvement with finetuning is diminished compared to the BCI and NHIE datasets. However, AnchorInv continues to outperform baseline methods, despite a smaller improvement margin. Breaking the score into base class and incremental class performance, we see that AnchorInv achieves the best incremental class performance while maintaining competitive base class performance. We also observe an improvement in performance in session 2 over session 1 for the NHIE dataset. This is because the incremental class introduced in session 2 appears to be relatively distinct from the remaining classes, as seen in Figure \ref{fig:visualization}. The high F1 score in the new class improves the average across all classes. Overall, the experiments on three datasets showcase the advantages of AnchorInv, especially when the number of base classes available is very limited.

\subsubsection{Qualitative Analysis of Inversion}
\begin{figure}[!h]
    \centering
    \includegraphics[width=0.9\linewidth]{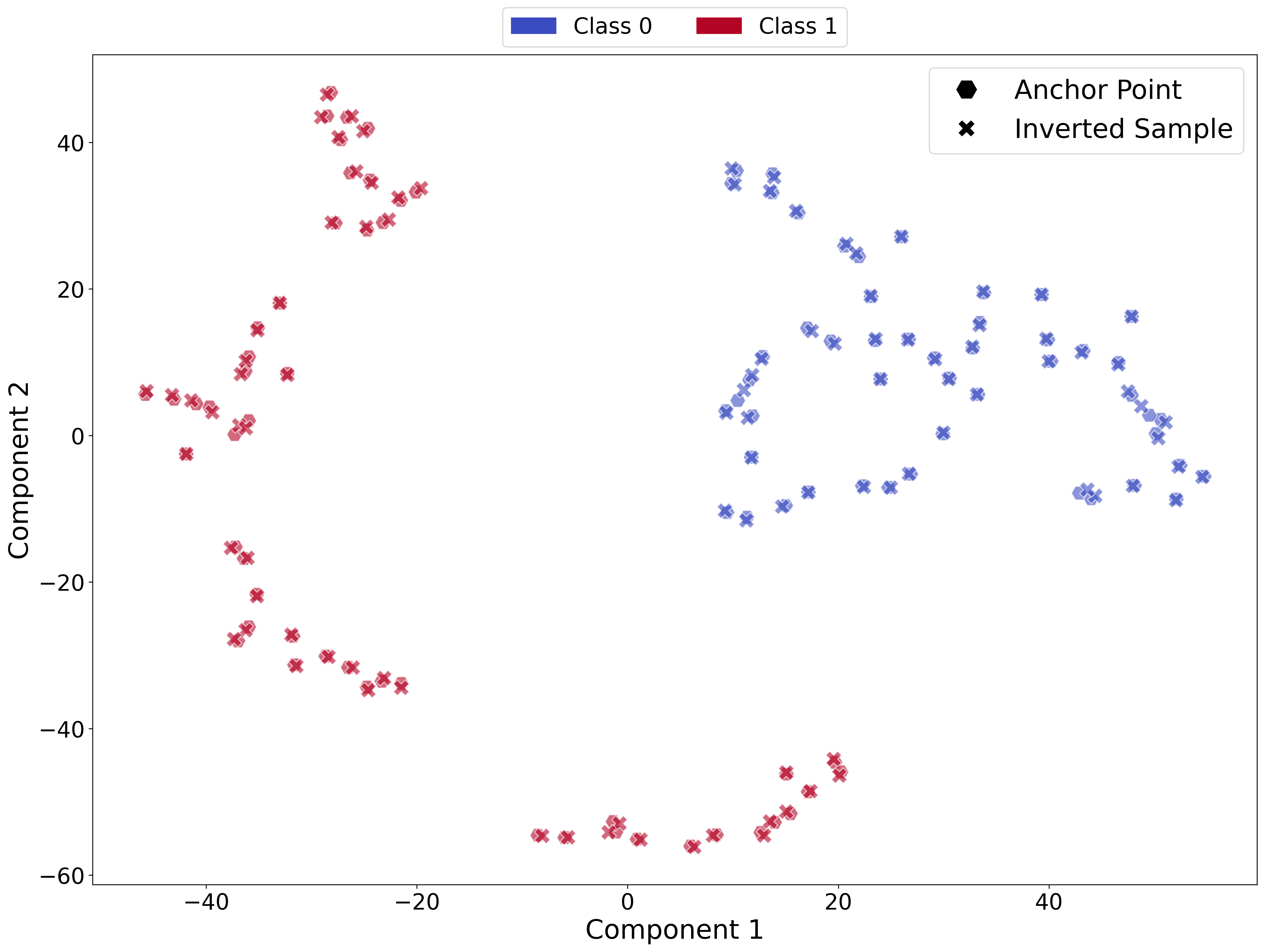}
    \caption{\textbf{t-SNE Visualization of the Anchor Set and the Inverted Samples in Feature Space}}
    \label{fig:invertvsrealfeature}
\end{figure}

Figure \ref{fig:invertvsrealfeature} visualizes the anchor set and the feature vector of the inverted replay set for the two base classes of the NHIE dataset. Note that the inverted feature vector and the anchor points overlap. This shows that the proposed model inversion algorithm can synthesize inverted samples that project to any target anchor point in the feature space. Appendix \ref{sec:realvsinvert} Figure \ref{fig:visualization} further compares the corresponding real and inverted \ac{EEG} segments from all four classes of the NHIE dataset. Despite sharing nearly identical feature vectors (mean absolute error between 0.06-0.08), the real and inverted samples are visually distinct. In particular, the inverted samples appear to be dominated by high-frequency fluctuations while the real samples observe more low-frequency fluctuations. Furthermore, high amplitude fluctuations tend to be observed towards the beginning of the inverted segments. By visual inspection, class 0 enjoys much higher amplitude fluctuations than class 4 in the real EEG segments, and the same trend can be observed for the inverted EEG segments.

\subsection{Ablation Studies}
\subsubsection{Selection of Anchor Points from Base Session}
We repeat AnchorInv experiments from Table \ref{tab:bci_results}, \ref{tab:nhie_results} and \ref{tab:grabm_results} and study how different anchor selection algorithms for the base classes impact the last session performance. Instead of random sampling, we select anchor points as feature vectors that are closest to class prototypes, denoted as ``Closest". We also apply $k$-means clustering to group each base class into five clusters and use the cluster centroids as anchor points, denoted as ``5 $k$-means Cluster". We also randomly sample 50 anchors from 30\%, 50\%, and 70\% of the feature vectors closest to each class prototype, denoted as ``Random Closest \{30, 50, 70\}\%". The selected anchor points guide the inversion of 50 samples for each base class in all experiments. As shown in Table \ref{tab:anchorselect}, there does not appear to be a distinct anchor selection algorithm that outperforms others in all datasets. In the case of a small number of base classes, BCI prefers anchor points that are closer to the prototype, while NHIE prefers more diverse anchor points. This could be due to the fact that BCI is trained and tested on the same subject, while NHIE is trained and tested on different subjects. The setup of the cross-subject train-test requires more diverse anchor points to fully represent the base class distributions, as illustrated through the t-SNE visualization in Appendix \ref{sec:tsne} Figures \ref{fig:bcitsne} and \ref{fig:nhietsne}. When a larger number of base classes is available, GRABMyo is less sensitive to the choice of anchor points than BCI and NHIE. Overall, clustering and random sampling are viable options for selecting anchor points, independent of the dataset. By Occam's razor principle, we prefer random sampling over clustering due to its simplicity.

\begin{table} 
\resizebox{\linewidth}{!}{
\begin{tabular}{llll}
\toprule \toprule
\textbf{Anchor Selection} & \textbf{BCI} & \textbf{NHIE} & \textbf{GRABMyo} \\
\midrule
Closest             & 38.65 $\pm$ 1.26 \, $\dagger$ & 62.63$\pm$2.38 \, $\dagger$ & 60.67$\pm$0.70 \\
5 $k$-means Cluster    & 36.97$\pm$1.28 & 66.05$\pm$2.90 & 60.70$\pm$0.68 \\
Random Closest 30\% & 38.36$\pm$1.34 \, $\dagger$ & 63.07$\pm$2.39 \, $\dagger$ & 60.67$\pm$0.71 $\dagger$ \\
Random Closest 50\% & 38.55$\pm$1.30 \, $\dagger$ & 64.03$\pm$2.23 \, $\dagger$ & 60.76$\pm$0.57 \\
Random Closest 70\% & 38.05$\pm$1.61 \, $\dagger$ & 64.21$\pm$2.33 \, $\dagger$ & 60.68$\pm$0.71 \\
Random Sample       & 37.04$\pm$1.24 & 66.03$\pm$2.73 & 60.78$\pm$0.66 \\
\bottomrule \bottomrule
\end{tabular}%
}
\caption{\textbf{Ablation on the Selection of Anchor Points and its Impact on Last Session Performance.} We report macro-F1 (\%) across all classes. $\dagger$ indicates results are statistically significantly different from the Random Sample with $p < 0.05$ using a two-sided Wilcoxon signed-rank test.}
\label{tab:anchorselect}
\end{table}

\subsubsection{Number of Shots for Incremental Training Sets}
We further study how the number of support samples available for each incremental class, $K$, impacts the performance of the last session. We vary $K$ between $\left\{ 1, 5, 10, 15, 20, 25, 30\right\}$. Figure \ref{fig:num_shot} shows that the performance stabilizes beyond 10 shots.

\subsubsection{Number of Anchor Points from Base Session}
\begin{figure}[!h]
    \centering
    \begin{subfigure}[b]{0.49\linewidth}
        \centering
        \includegraphics[width=\linewidth]{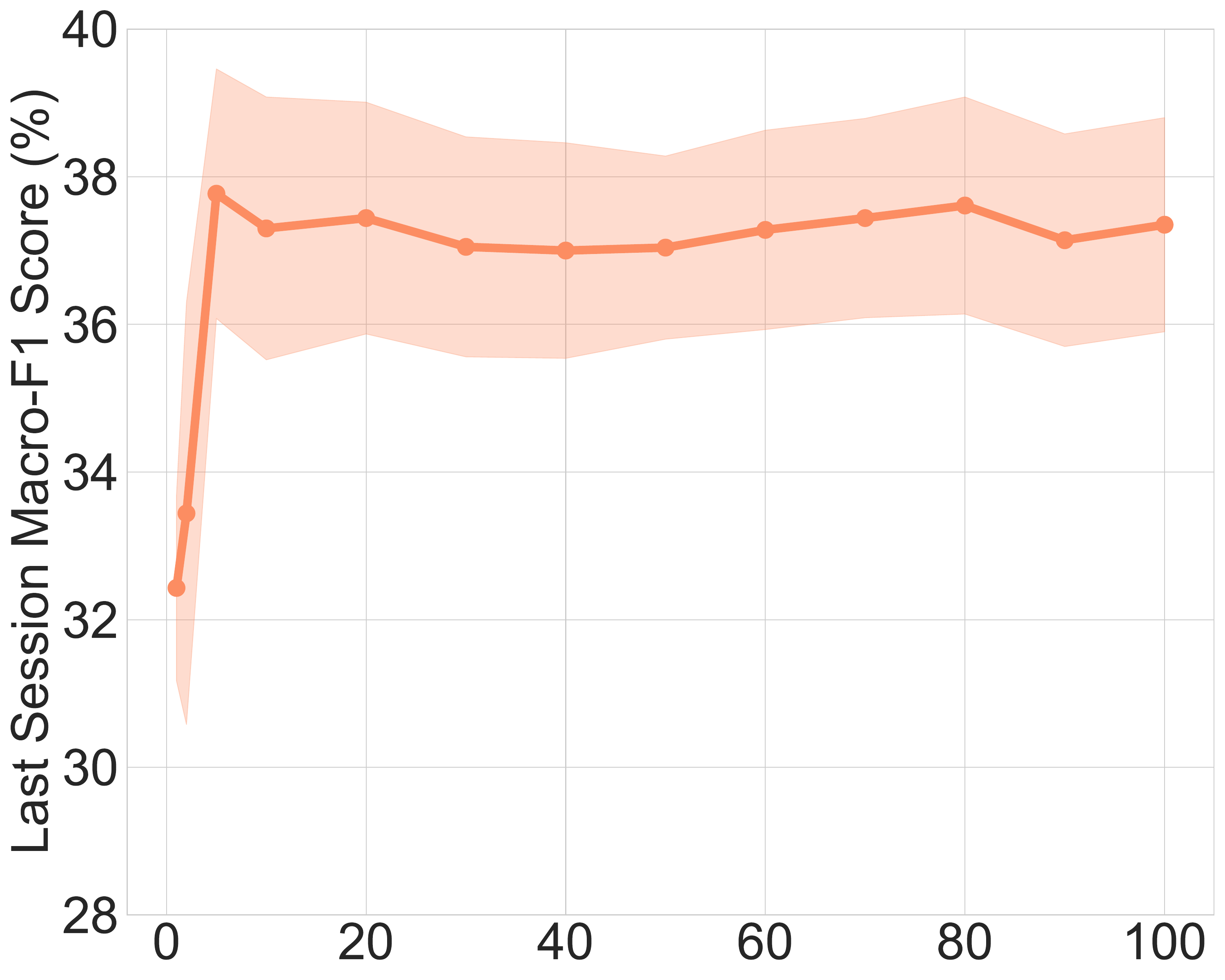}
        \caption{$M$}
        \label{fig:num_invert}
    \end{subfigure}
    \hfill
    \begin{subfigure}[b]{0.49\linewidth}
        \centering
        \includegraphics[width=\linewidth]{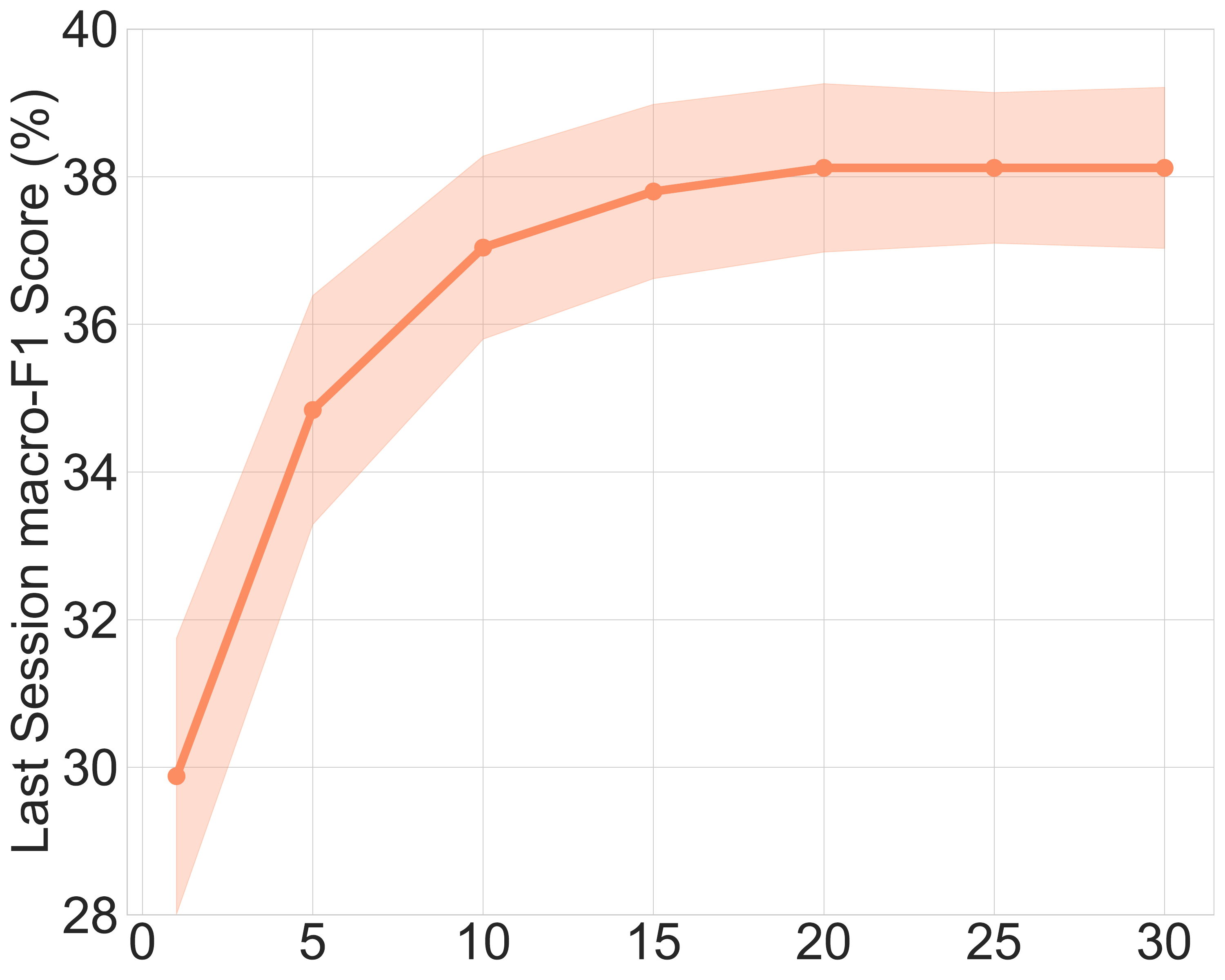}
        \caption{$K$}
        \label{fig:num_shot}
    \end{subfigure}
    \caption{\textbf{AnchorInv Ablations.} We compare the last session's performance on the BCI dataset under varying parameter settings. (a) $M$ Number of anchor points per base class (b) $K$ Number of training samples per incremental class}
    \label{fig:combined_figures}
\end{figure}

We further study how the number of anchor points stored from the base session impacts the last session performance on BCI. We randomly select $P = \left\{ 1, 2, 5, 10, 20, 30, 40, 50, 60, 70, 80, 90, 100 \right\} $ anchor points and invert the same number of samples in session 1. Figure \ref{fig:num_invert} shows that as the number of inverted samples increases, the macro-F1 score of the last session increases and stabilizes.

\subsubsection{Additional Ablations}
We further investigate the variation of the number of base session classes for GRABMyo and the comparison with real anchor samples as an upper bound in Section \ref{sec:additionalablations} of the Supplementary Materials.

\section{Conclusion}
In this study, we investigate \ac{FSCIL} for physiological time series and note that existing methods rely heavily on a large number of base classes or violate data sharing and privacy concerns between sessions. To this end, we investigate model inversion guided by anchor points to retrieve diverse representations for previously seen classes. We further investigate how the selection of anchor points impacts incremental session performance. Experiments across multiple trials show that feature space-guided inversion helps to learn new classes better while maintaining a more faithful representation of previously seen classes.

\newpage

\bibliography{aaai25}

\newpage
\section*{Reproducibility Checklist}
This paper:
\begin{itemize}
    \item Includes a conceptual outline and/or pseudocode description of AI methods introduced \textbf{(yes)}
    \item Clearly delineates statements that are opinions, hypothesis, and speculations from objective facts and results \textbf{(yes)}
    \item Provides well marked pedagogical references for less-familiare readers to gain background necessary to replicate the paper \textbf{(yes)}
    \item Does this paper make theoretical contributions? \textbf{(yes)}
\end{itemize}

\noindent All assumptions and restrictions are stated clearly and formally. \textbf{(yes)}
\begin{itemize}
    \item All novel claims are stated formally (e.g., in theorem statements). \textbf{(yes)}
    \item Proof sketches or intuitions are given for complex and/or novel results. \textbf{(yes)}
    \item Appropriate citations to theoretical tools used are given. \textbf{(yes)}
    \item All theoretical claims are demonstrated empirically to hold. \textbf{(yes)}
    \item All experimental code used to eliminate or disprove claims is included. \textbf{(yes)}
    \item Proofs of all novel claims are included. \textbf{(yes)}
\end{itemize}

\noindent Does this paper rely on one or more datasets? \textbf{(yes)}

\begin{itemize}
    \item A motivation is given for why the experiments are conducted on the selected datasets \textbf{(yes)}
    \item All novel datasets introduced in this paper are included in a data appendix. \textbf{(yes)}
    \item All novel datasets introduced in this paper will be made publicly available upon publication of the paper with a license that allows free usage for research purposes. \textbf{(yes)}
    \item All datasets drawn from the existing literature (potentially including authors’ own previously published work) are accompanied by appropriate citations. \textbf{(yes)}
    \item All datasets drawn from the existing literature (potentially including authors’ own previously published work) are publicly available. \textbf{(yes)}
    \item All datasets that are not publicly available are described in detail, with explanation why publicly available alternatives are not scientifically satisficing. \textbf{(yes)}

\end{itemize}

\noindent Does this paper include computational experiments? \textbf{(yes)}

\begin{itemize}
    \item Any code required for pre-processing data is included in the appendix. \textbf{(yes)}
    \item All source code required for conducting and analyzing the experiments is included in a code appendix. \textbf{(yes)}
    \item All source code required for conducting and analyzing the experiments will be made publicly available upon publication of the paper with a license that allows free usage for research purposes. \textbf{(yes)}
    \item All source code implementing new methods have comments detailing the implementation, with references to the paper where each step comes from. \textbf{(yes)}
    \item If an algorithm depends on randomness, then the method used for setting seeds is described in a way sufficient to allow replication of results. \textbf{(yes)}
    \item This paper specifies the computing infrastructure used for running experiments (hardware and software), including GPU/CPU models; amount of memory; operating system; names and versions of relevant software libraries and frameworks. \textbf{(yes)}
    \item This paper formally describes evaluation metrics used and explains the motivation for choosing these metrics. \textbf{(yes)}
    \item This paper states the number of algorithm runs used to compute each reported result. \textbf{(yes)}
    \item The significance of any improvement or decrease in performance is judged using appropriate statistical tests (e.g., Wilcoxon signed-rank). (\textbf{yes})
    \item Analysis of experiments goes beyond single-dimensional summaries of performance (e.g., average; median) to include measures of variation, confidence, or other distributional information. \textbf{(yes)}
    \item This paper lists all final (hyper-)parameters used for each model/algorithm in the paper’s experiments. \textbf{(yes)}
    \item This paper states the number and range of values tried per (hyper-) parameter during development of the paper, along with the criterion used for selecting the final parameter setting. \textbf{(yes)}

\end{itemize}

\newpage
\appendix

\section{Dataset Preprocessing} \label{sec:preprocessing}
\subsection{Brain-Computer Interface Motor-Imagery (BCI)}
A total of two sessions of data were collected on different days for 9 subjects. We follow the train-test split of the original dataset, where one session from each subject is used for training and the other session is used for testing. The signals were sampled at 250 Hz with the 10-20 international system. The dataset has been preprocessed with a notch filter at 50 Hz and a bandpass filter between 0.5 and 100 Hz to suppress noise. We further apply Z-score standardization to both the training and test set, using training data statistics:
\begin{equation} \label{eq:z_standardization}
    \mathbf{x} = \frac{\mathbf{x}_f - \mu}{\sigma}
\end{equation}
where $\mathbf{x}_f$ and $\mathbf{x}$ denotes the \ac{EEG} signal after filtering and after standardization respectively. $\mu$ and $\sigma$ denote the mean and standard deviation of training data. After all preprocessing, each sample is four seconds in length with 22 channels, resulting in an input shape of (22, 1000).
\subsection{Grading of Neonatal Hypoxic-Ischemic Encephalopathy (NHIE)}
A total of 169 one-hour recordings are available from 53 neonates. Each neonate has a minimum of one and a maximum of five one-hour recordings, with varying combinations of grades. We perform a stratified subject-independent train-test split in an 80-20 ratio, resulting in a training and test set that contains different subjects but has the same proportion for each grade. The signals were collected using the 10-20 international system across two different \ac{EEG} devices, with sample rates of 256 Hz and 200 Hz. Following the preliminary analysis in \cite{o2023neonatal}, we convert to a bipolar montage with pairs: F3–C3, T3–C3, O1–C3, C3–Cz, Cz–C4, F4–C4, T4–C4, and O2–C4. The signals are processed with a bandpass filter between 0.5 and 30 Hz and downsampled to 64 Hz. Per-channel mean and standardization deviation were calculated in the training set and applied to both the training set and the test set following Equation \ref{eq:z_standardization}. Each one-hour recording is further divided into 60-second segments, with 50 \% overlap, resulting in a total of 119 segments per recording. After all preprocessing, each sample is of input shape (8, 3840).

\subsection{Gesture Recognition (GRABMyo)}
Data is available from 43 participants across three different sessions. Similarly to the train-test split from the BCI dataset, we select the data from session 1 and session 2 to form the training set, and the data from session 3 form the test set. The signals were sampled at 2048 Hz with 16 forearm and 12 wrist channels. The dataset has been preprocessed with a notch filter at 60 Hz and a fourth-order Butterworth bandpass filter between 10 Hz and 500 Hz \cite{pradhan2022multi}. Following \cite{qiao2024class}, we downsample the signal to 64 Hz. Z-score standardization is applied similarly to training and test sets following Equation \ref{eq:z_standardization}. The final samples are five seconds in length, resulting in an input shape of (28, 1280).

\section{Convolution Module Architecture} \label{sec:conv_module_architecture}

\begin{table}[h]
\centering
\begin{tabular}{lcccc}
\toprule
\toprule
Layer          & In       & Out      & Kernel     & Stride  \\ \midrule
Temporal Conv  & $1$        & $40$        & $(1, 25)$    & $(1, 1)$  \\
Spatial Conv   & $40$        & $40$        & $(22, 1)$    & $(1, 1)$  \\
Avg Pooling    & $40$        & $40$        & $(1, 75)$    & $(1, 15)$ \\
Rearrange      & \multicolumn{4}{c}{$(40, 1, 61) \rightarrow (61, 40)$}  \\
\bottomrule
\bottomrule
\end{tabular}
\caption{Convolution Module for BCI Dataset}
\end{table}

\begin{table}[h]
\centering
\begin{tabular}{lcccc}
\toprule
\toprule
Layer          & In       & Out      & Kernel     & Stride  \\ \midrule
Temporal Conv  & $1$        & $40$        & $(1, 64)$    & $(1, 4)$  \\
Spatial Conv   & $40$        & $40$        & $(8, 1)$    & $(1, 1)$  \\
Avg Pooling    & $40$        & $40$        & $(1, 75)$    & $(1, 15)$ \\
Rearrange      & \multicolumn{4}{c}{$(40, 1, 59) \rightarrow (59, 40)$}  \\
\bottomrule
\bottomrule
\end{tabular}
\caption{Convolution Module for NHIE Dataset}
\end{table}

\begin{table}[h]
\centering
\begin{tabular}{lcccc}
\toprule
\toprule
Layer          & In       & Out      & Kernel     & Stride  \\ \midrule
Temporal Conv  & $1$        & $40$        & $(1, 64)$    & $(1, 1)$  \\
Spatial Conv   & $40$        & $40$        & $(28, 1)$    & $(1, 1)$  \\
Avg Pooling    & $40$        & $40$        & $(1, 75)$    & $(1, 20)$ \\
Rearrange      & \multicolumn{4}{c}{$(40, 1, 58) \rightarrow (58, 40)$}  \\
\bottomrule
\bottomrule
\end{tabular}
\caption{Convolution Module for GRABMyo Dataset}
\end{table}

\section{Model Training and Adaptation} \label{sec:trainadapt}
All experiments are performed using PyTorch on NVIDIA GeForce RTX 3090 GPUs with 24GB memory. A set seed of 5 is used for all experiments.
\subsection{Base Session Training} \label{sec:basesessiontrain}
In the base session, we train the embedding network with loss of cross-entropy using the Adam optimizer with default momentum parameters \(\beta_1 = 0.9\), \(\beta_2 = 0.999\). In BCI, we train for 1500 epochs with a learning rate of \(2 \times 10^{-4}\) and a batch size of 72 in the base session. In NHIE, we train for 2000 epochs with a learning rate of \(1 \times 10^{-5}\), a batch size of 256, and a weighted loss inversely proportional to the number of samples per class. In GRABMyo, we train for 2000 epochs with a learning rate of \(5 \times 10^{-5}\) and a batch size of 256. Across all datasets, we select $M=50$ anchor points for each base class with $\mu$ as the random sampling algorithm. ProtoNet, TEEN, DeepDream, DeepInv, and AnchorInv adopt a metric-based classifier, as described in Equations \ref{eq:protopred3}. They share the same base session training and use the same checkpoint for incremental adaptation. Meanwhile, Finetune and NC-FSCIL adopt MLP as a classifier and are trained via CrossEntropy and DotRegression loss in the base session, respectively. Therefore, it is expected that the base session performance will be the same for ProtoNet, TEEN, DeepDream, DeepInv, and AnchorInv but differ from that of Finetune and NC-FSCIL.

\subsection{Incremental Session Adaptation} \label{sec:incrementalsessionadapt}
In the incremental session, we initialize all inverted samples with a standard normal distribution with zero mean and unit variance. To guide inversion, we randomly select 50 anchors for each base class and save all 10 feature vectors for each incremental class. We invert 50 samples for each base class and 10 samples for each incremental class. During inversion, we optimize using the \ac{MAE} loss and Adam optimizer with \(1 \times 10^{-2}\) learning rate for 4000, 4000, and 2000 iterations across BCI, NHIE, and GRABMyo datasets respectively. With the inverted samples and the few-shot support samples for the new class, we finetune the last layer of the backbone and the classifier using the Adam optimizer. Only the incremental class weights of the classifier are finetuned, the classifier weights of classes from prior sessions are frozen. In BCI, we finetune for 1000-1550 iterations with a learning rate of \(2 \times 10^{-5}\). In NHIE, we finetune for 700-1250 iterations with a learning rate of \(1 \times 10^{-5}\). In GRABMyo, we initialize the prototype vector for the new class as the average embedding across the training samples, then we finetune for 0-1300 iterations with a learning rate of \(5 \times 10^{-6}\).

\section{Baseline Method Hyperparameters}
For consistency, we adopt the same backbone architecture throughout all experiments for each dataset. All algorithms are trained using Adam optimizer with default momentum parameters \(\beta_1 = 0.9\), \(\beta_2 = 0.999\). To choose the number of iterations to finetune in incremental sessions, we run 10 trials and select the best-performing iteration (at 50 iteration resolution) based on test set performance.
\subsubsection{Finetune} 
We adopt the fully connected classifier from EEG-Conformer \cite{song2023eeg}, which consists of two fully connected layers with 256 and 32 neurons respectively. We follow the same base session training hyperparameters as described in Section \ref{sec:basesessiontrain}. In the incremental sessions, we finetune the model using incremental training samples at a learning rate of \(2 \times 10^{-5}\), \(1 \times 10^{-5}\), \(5 \times 10^{-6}\) for 0-150 iterations on BCI, NHIE and GRABMyo respectively. 
\subsubsection{Neural Collapse}
We adopt the same projection layer and ETF classifier as described in the original work \cite{yang2023neural}. We apply an adaptive pool to reduce the feature vector dimension to 512 before the projection layer. In the base session, we train the backbone and projection layer jointly following Section \ref{sec:basesessiontrain}, but we adopt the dot-regression loss \cite{yang2023neural} with weight 10. In the incremental sessions, we finetune the projection layer for 150-1000 iterations at a learning rate of \(2 \times 10^{-5}\), \(1 \times 10^{-5}\), \(5 \times 10^{-6}\) for BCI, NHIE, and GRABMyo respectively.
\subsubsection{ProtoNet}
In the base session, we train the backbone with $T=16$ following Section \ref{sec:basesessiontrain}. In the incremental sessions, we freeze the backbone and calculate class prototypes following Eq. \ref{eq:proto}.
\subsubsection{TEEN}
In the base session, we train the backbone with $T=16$ following Section \ref{sec:basesessiontrain}. In the incremental sessions, we calculate class prototypes following Eq. \ref{eq:proto} and calibrate the prototypes with $\tau=32$ and $\alpha=0.5$.
\subsubsection{DeepDream and DeepInv}
In the base session, we train the backbone with $T=16$ following Section \ref{sec:basesessiontrain}. We follow the same hyperparameters for inversion as described in Section \ref{sec:incrementalsessionadapt}, but replace the MAE loss with cross-entropy loss. In addition, we follow \cite{yin2020dreaming} and set the weights for the $l$2, total variation, and feature distribution regularization terms so that all loss terms are of the same magnitude. We finetune the last layer of the backbone for 0-3450 iterations with a learning rate of \(2 \times 10^{-5}\), \(1 \times 10^{-5}\), \(5 \times 10^{-6}\) for BCI, NHIE, and GRABMyo respectively.

\section{Additional Ablation Studies} \label{sec:additionalablations}
\subsection{Varying Number of Base Session Classes}
We vary the number of base classes for GRAMyo and keep the same incremental classes as Table \ref{tab:grabm_results}. From Table \ref{tab:baseclassablation}, we observe that AnchorInv demonstrates a substantial improvement over baselines in the low-base class regime.
\begin{table*}[!h] 
\begin{subtable}[t]{\textwidth}
\subcaption{\textbf{GRABMyo 6 base classes 1-way-10-shot \ac{FSCIL} across 10 trials (\%)}}
\resizebox{1\textwidth}{!}{
\begin{tabular}{ccccccccc}
\toprule
\toprule
\multicolumn{1}{c}{\multirow{2}{*}{\textbf{Method}}} & \multicolumn{7}{c}{\textbf{Macro-F1 (All Classes)}} \\
\cmidrule(lr){2-8}
\multicolumn{1}{c}{}                           & \textbf{Base Session} & \textbf{Session 1} & \textbf{Session 2} & \textbf{Session 3}   & \textbf{Session 4}   & \textbf{Session 5} & \textbf{Session 6}   \\
\midrule

      Finetune &        91.45 & 70.04 $\pm$ 0.74 & 57.23 $\pm$ 0.72 & 48.18 $\pm$ 1.21 & 41.06 $\pm$ 1.23 & 35.57 $\pm$ 1.53 & 29.63 $\pm$ 1.19 \\
      NC-FSCIL &        91.55 & 81.91 $\pm$ 1.46 & 74.04 $\pm$ 1.39 & 65.41 $\pm$ 1.65 & 59.03 $\pm$ 1.29 & 54.04 $\pm$ 1.21 & 47.88 $\pm$ 1.05 \\
      ProtoNet &        91.13 & 84.15 $\pm$ 0.65 & 75.83 $\pm$ 0.79 & 68.36 $\pm$ 0.62 & 61.07 $\pm$ 1.05 & 58.33 $\pm$ 0.83 & 54.80 $\pm$ 1.04 \\
      TEEN &        91.13 & 82.52 $\pm$ 1.38 & 74.54 $\pm$ 1.42 &  67.04 $\pm$ 1.00 & 59.25 $\pm$ 0.76 & 57.29 $\pm$ 0.80 & 53.85 $\pm$ 1.01 \\
      DeepDream &        91.13 & \textbf{85.75 $\pm$ 0.99} & 76.98 $\pm$ 1.16 & 66.96 $\pm$ 1.04 & 59.51 $\pm$ 1.68 & 54.78 $\pm$ 2.74 & 52.32 $\pm$ 2.36 \\
      DeepInv &        91.13 & 85.60 $\pm$ 1.15 & 76.84 $\pm$ 0.99 & 67.66 $\pm$ 1.54 & 59.81 $\pm$ 1.78 & 55.98 $\pm$ 1.82 & 50.80 $\pm$ 1.85 \\
      AnchorInv &        91.13 & 85.54 $\pm$ 0.70 & \textbf{77.70 $\pm$ 0.46} & \textbf{70.13 $\pm$ 0.54} & \textbf{63.14 $\pm$ 1.05} & \textbf{60.63 $\pm$ 0.84} & \textbf{56.63 $\pm$ 0.60} \\
\bottomrule
\bottomrule
\end{tabular}%
}
\end{subtable}

\begin{subtable}[t]{\textwidth}
\subcaption{\textbf{GRABMyo 4 base classes 1-way-10-shot \ac{FSCIL} across 10 trials (\%)}}
\resizebox{1\textwidth}{!}{
\begin{tabular}{ccccccccc}
\toprule
\toprule
\multicolumn{1}{c}{\multirow{2}{*}{\textbf{Method}}} & \multicolumn{7}{c}{\textbf{Macro-F1 (All Classes)}} \\
\cmidrule(lr){2-8}
\multicolumn{1}{c}{}                           & \textbf{Base Session} & \textbf{Session 1} & \textbf{Session 2} & \textbf{Session 3}   & \textbf{Session 4}   & \textbf{Session 5} & \textbf{Session 6}   \\
\midrule

Finetune &        91.13 & 60.69 $\pm$ 1.94 & 45.04 $\pm$ 1.88 & 33.86 $\pm$ 4.04 & 26.70 $\pm$ 2.70 & 22.43 $\pm$ 3.13 & 17.54 $\pm$ 3.48 \\
      NC-FSCIL &        91.14 & 76.68 $\pm$ 1.33 & 62.16 $\pm$ 1.95 & 52.13 $\pm$ 1.57 & 45.58 $\pm$ 1.38 & 40.94 $\pm$ 1.63 & 21.92 $\pm$ 0.25 \\
      ProtoNet &        91.22 & 77.39 $\pm$ 0.63 & 66.23 $\pm$ 0.62 & 56.88 $\pm$ 0.90 & 49.03 $\pm$ 1.08 & 46.64 $\pm$ 1.16 & 42.99 $\pm$ 1.64 \\
      TEEN &        91.22 & 76.09 $\pm$ 1.09 & 64.72 $\pm$ 0.79 & 55.94 $\pm$ 0.95 & 48.69 $\pm$ 1.38 & 47.08 $\pm$ 1.41 & 43.40 $\pm$ 2.15 \\
      DeepDream &        91.22 & 80.28 $\pm$ 0.83 & 66.16 $\pm$ 3.58 & 55.61 $\pm$ 3.07 & 46.08 $\pm$ 2.64 & 42.32 $\pm$ 2.42 & 37.62 $\pm$ 2.53 \\
      DeepInv &        91.22 & 80.33 $\pm$ 0.81 & 67.20 $\pm$ 2.78 & 56.56 $\pm$ 1.10 & 47.38 $\pm$ 1.44 & 42.47 $\pm$ 1.45 & 39.67 $\pm$ 2.61 \\
      AnchorInv &        91.22 & \textbf{80.72 $\pm$ 0.77} & \textbf{68.59 $\pm$ 1.44} & \textbf{59.51 $\pm$ 1.01} & \textbf{49.70 $\pm$ 2.32} & \textbf{48.60 $\pm$ 1.51} & \textbf{44.76 $\pm$ 1.23} \\
\bottomrule
\bottomrule
\end{tabular}%
}
\end{subtable}
\caption{\textbf{Comparison of AnchorInv against Baseline}. Across all incremental sessions, AnchorInv is statistically significantly different from all baseline methods with $p < 0.05$ using a two-sided Wilcoxon signed-rank test.}
\label{tab:baseclassablation}
\

\end{table*}

\subsection{Comparison with Real Anchor Samples}
In Table \ref{tab:realanchor}, we conduct further experiments to compare the fine-tuning using real samples as opposed to inverted samples. In RealReplay, we randomly select and store 50 real samples from the base session and all samples from incremental sessions. We use the stored samples to regularize prior knowledge during fine-tuning with new samples. Although it violates privacy limitations, it serves as an upper bound for the proposed method. We observe that AnchorInv's performance is very similar to that of RealReplay. This indicates that the proposed inversion method synthesizes samples that contain information similar to that in the real samples to help maintain a faithful representation of previously seen classes while protecting privacy. 

\begin{table}[]
\centering
\begin{tabular}{cccc}
\toprule
\toprule
\multicolumn{1}{c}{\multirow{2}{*}{\textbf{Method}}} & \multicolumn{3}{c}{\textbf{Macro-F1 (All Classes)}} \\
\cmidrule(lr){2-4}
\multicolumn{1}{c}{}                           & \textbf{Base Session} & \textbf{Session 1} & \textbf{Session 2} \\
\midrule

Finetune &        78.81 & 45.88 $\pm$ 2.25 & 31.57 $\pm$ 2.47 \\
    NC-FSCIL &        78.39 &  48.86 $\pm$ 1.83 & 35.43 $\pm$ 1.26 \\
ProtoNet &        77.17 & 45.31 $\pm$ 0.78 & 30.99 $\pm$ 1.19 \\
TEEN &        77.17 & 46.33 $\pm$ 0.95 & 32.84 $\pm$ 1.36 \\
DeepDream &        77.17 & 51.12 $\pm$ 1.58 & 33.92 $\pm$ 1.26 \\
DeepInv &        77.17 & 50.87 $\pm$ 0.99 & 34.23 $\pm$ 1.92 \\
AnchorInv &        77.17 & 51.26 $\pm$ 0.93 & 37.81 $\pm$ 1.31 \\
RealReplay &        77.17 & \textbf{51.27 $\pm$ 0.94} & \textbf{37.87 $\pm$ 1.42} \\
\bottomrule
\bottomrule
\end{tabular}
\caption{\textbf{Comparison with Real Anchor Samples.} BCI 1-way-10-shot \ac{FSCIL} across 10 trials (\%)} 
\label{tab:realanchor}
\end{table}

\subsection{Selection of Base Classes}
We train with all combinations of base classes and use the remaining classes as incremental classes. The header of each column in Table \ref{tab:basecombinations} indicates the base classes. We observe that AnchorInv consistently outperforms all baselines across all scenarios.

\begin{table*}[]
\centering
\begin{tabular}{cccccc}
\toprule
\toprule
\multicolumn{1}{c}{\multirow{2}{*}{\textbf{Method}}} & \multicolumn{5}{c}{\textbf{Macro-F1 (All Classes)}} \\
\cmidrule(lr){2-6}
\multicolumn{1}{c}{} & \textbf{0\&2} & \textbf{1\&2} & \textbf{1\&3} & \textbf{0\&3} & \textbf{2\&3} \\
\midrule

Finetune     & 25.03$\pm$2.30 & 27.21$\pm$1.37 & 30.99$\pm$1.41 & 27.43$\pm$1.98 & 25.39$\pm$2.30 \\
NC-FSCIL     & 31.90$\pm$1.02 & 32.41$\pm$1.32 & 33.21$\pm$1.11 & 34.64$\pm$1.10 & 30.94$\pm$1.18 \\
ProtoNet     & 33.99$\pm$0.81 & 33.30$\pm$1.54 & 33.84$\pm$1.72 & 36.12$\pm$1.64 & 33.99$\pm$0.81 \\
TEEN         & 35.81$\pm$0.79 & 34.62$\pm$1.82 & 35.36$\pm$1.94 & 37.92$\pm$2.07 & 35.81$\pm$0.79 \\
DeepDream    & 33.74$\pm$1.59 & 35.12$\pm$1.35 & 35.27$\pm$2.29 & 36.49$\pm$1.65 & 33.74$\pm$1.59 \\
DeepInv      & 35.33$\pm$1.39 & 35.71$\pm$1.77 & 36.11$\pm$2.46 & 37.58$\pm$1.53 & 35.33$\pm$1.39 \\
AnchorInv    & 38.63$\pm$1.11 & 38.28$\pm$1.34 & 40.85$\pm$1.25 & 41.21$\pm$1.15 & 38.63$\pm$1.10 \\

\bottomrule
\bottomrule
\end{tabular}
\caption{\textbf{Ablation on the Selection of Base Classes.} BCI 1-way-10-shot \ac{FSCIL} across 10 trials (\%)} 
\label{tab:basecombinations}
\end{table*}

\section{Visualization of Real and Inverted Samples} \label{sec:realvsinvert}
The inverted samples are initialized as a zero matrix and updated over 4000 iterations. The real and inverted pairs share the same anchor point in the feature space.
\begin{figure}
    \centering
    \includegraphics[width=\linewidth]{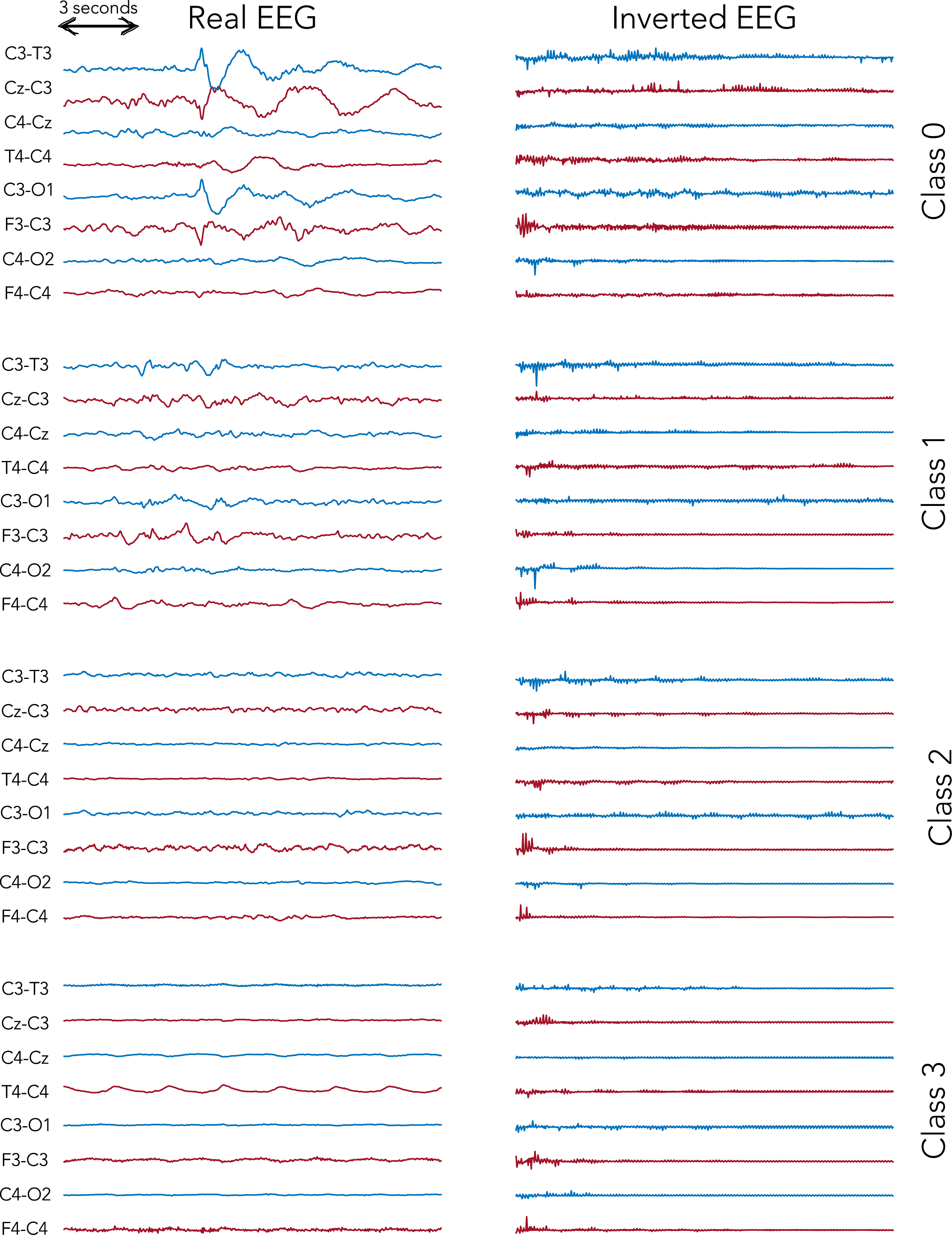}
    \caption{\textbf{Visualization of EEG Samples from NHIE}} 
    \label{fig:visualization}
\end{figure}

\section{Anchor Selection from Base Class Feature Distribution} \label{sec:tsne}

\begin{figure}
    \centering
    \begin{subfigure}[b]{\linewidth}
        \centering
        \includegraphics[width=\linewidth]{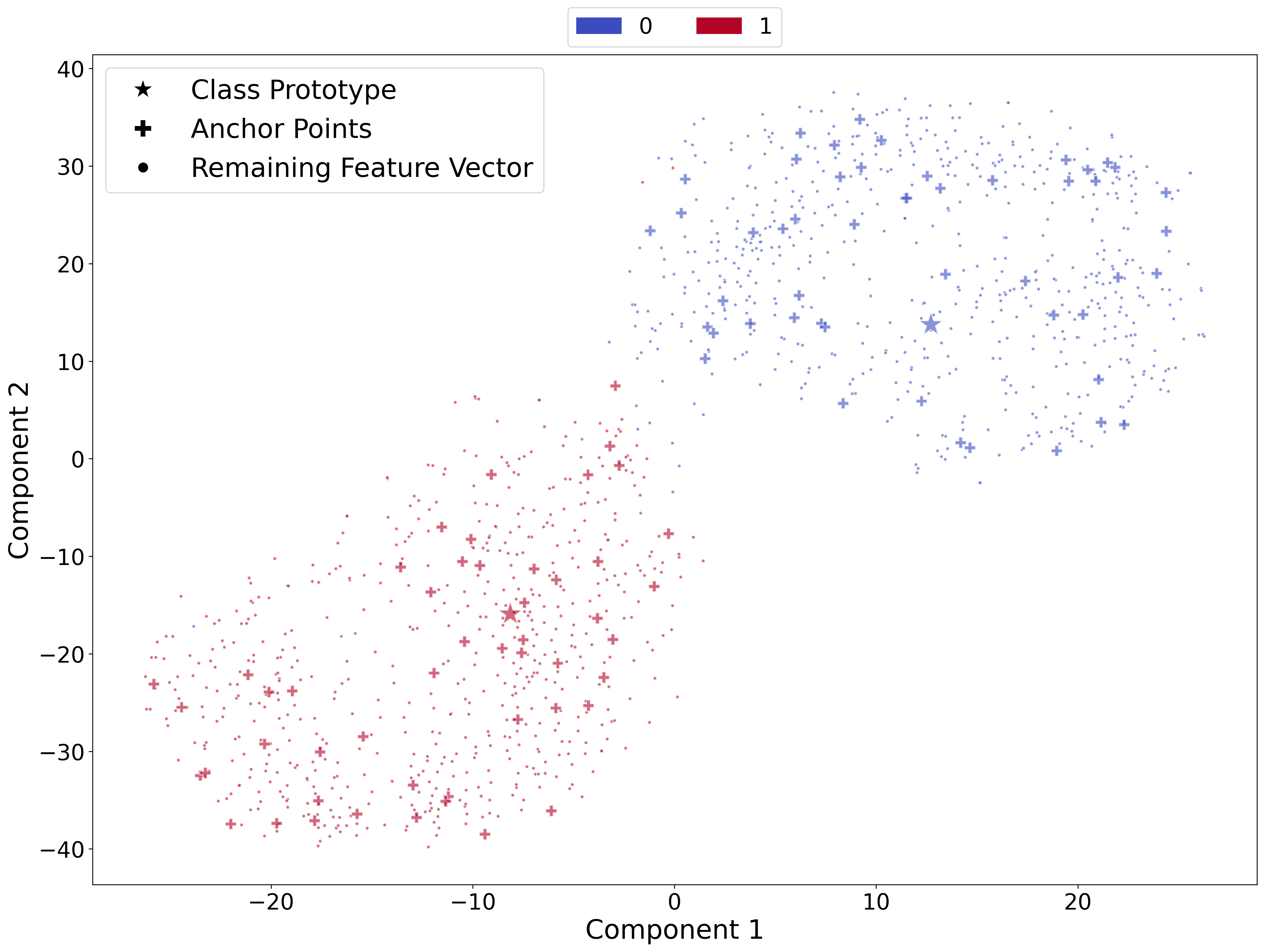}
        \caption{Anchor points selected as ``Random Sample"}
        \label{fig:closestbci}
    \end{subfigure}
    \hfill
    \begin{subfigure}[b]{\linewidth}
        \centering
        \includegraphics[width=\linewidth]{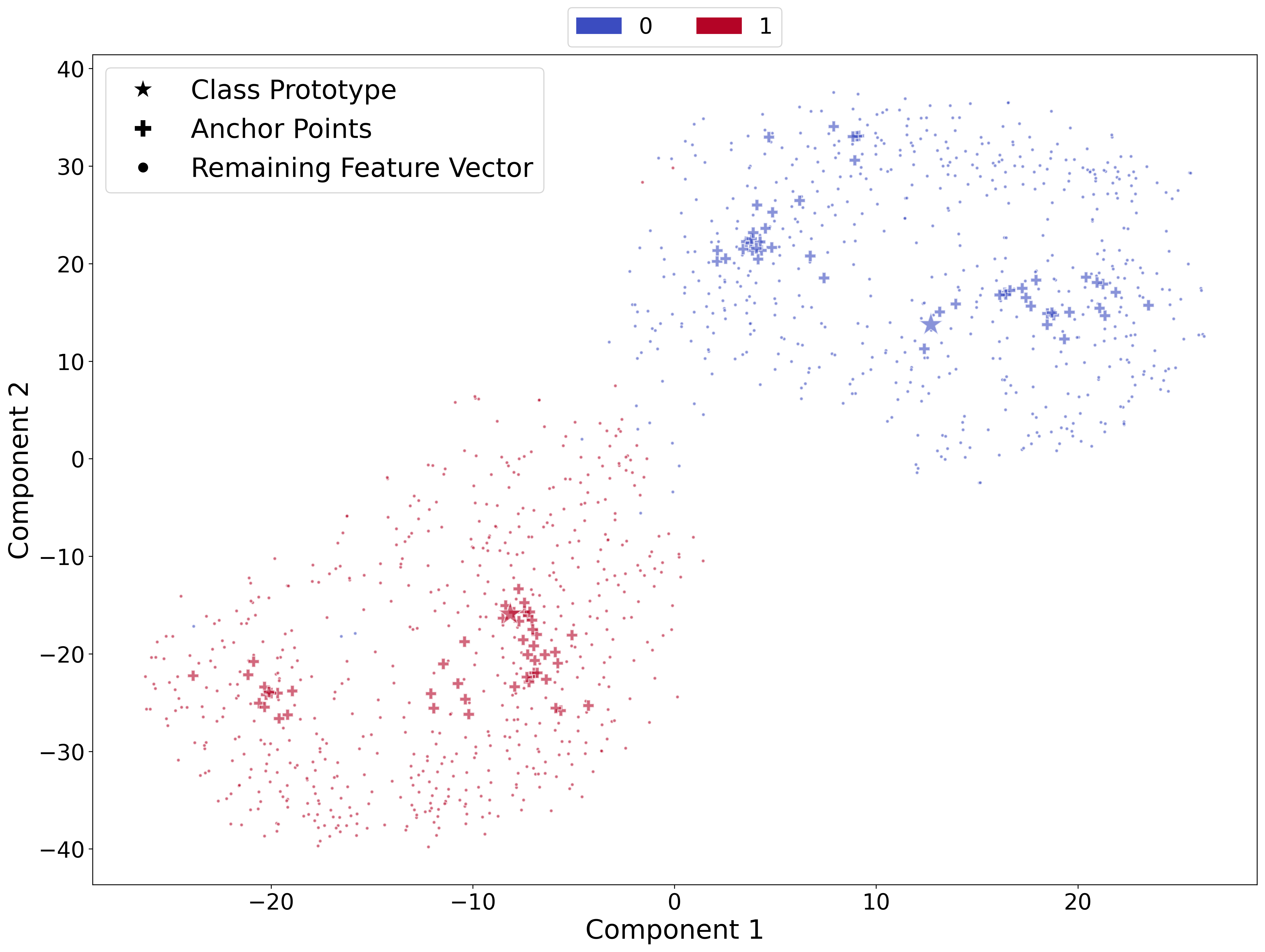}
        \caption{Anchor points selected through ``Closest" to prototype}
        \label{fig:randombci}
    \end{subfigure}
    \caption{\textbf{Visualization of Base Class Feature Space Distribution for the BCI Dataset}}
    \label{fig:bcitsne}
\end{figure}

\begin{figure}
    \centering
    \begin{subfigure}[b]{\linewidth}
        \centering
        \includegraphics[width=\linewidth]{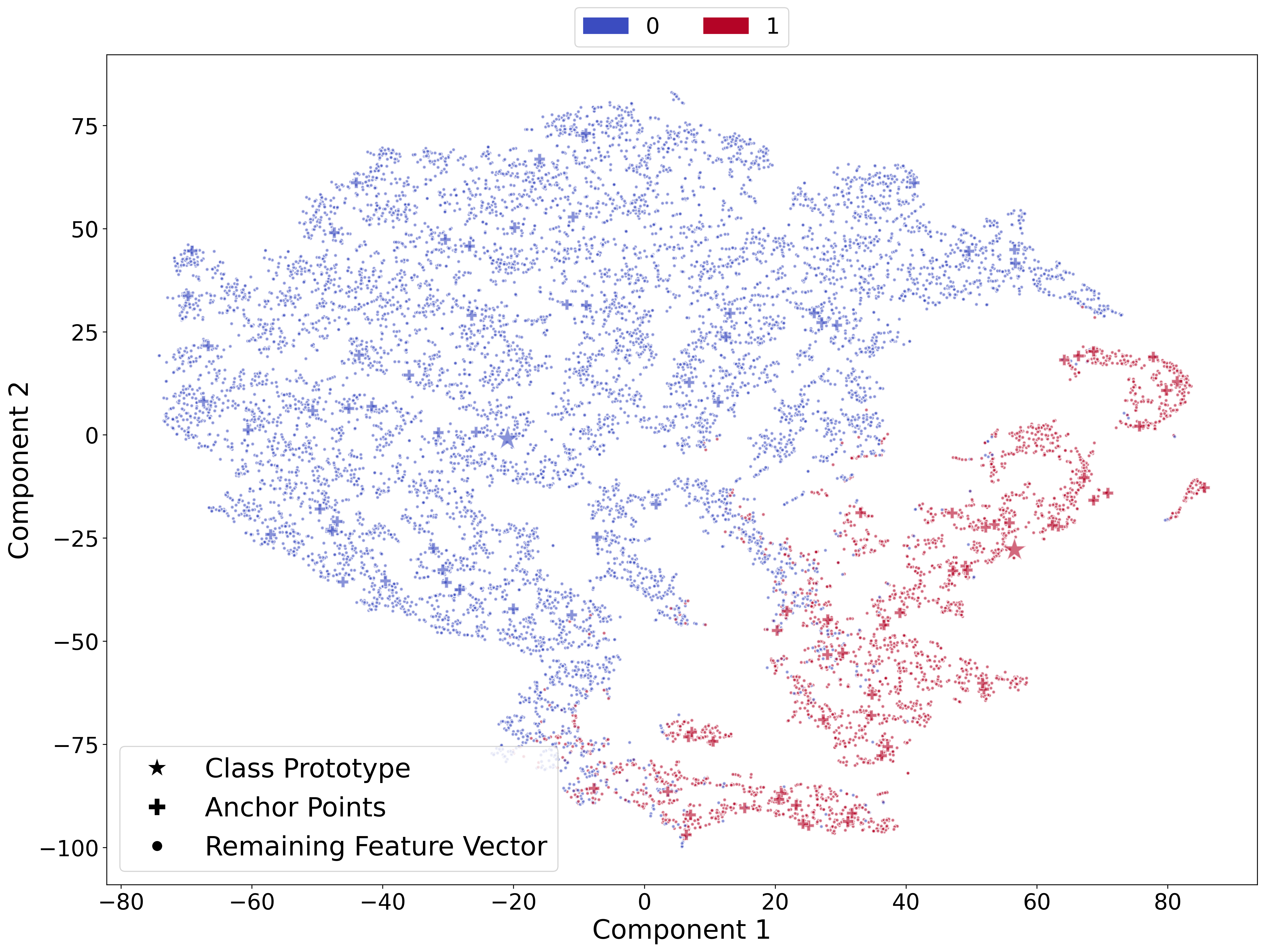}
        \caption{Anchor points selected as ``Random Sample"}
        \label{fig:closestnhie}
    \end{subfigure}
    \hfill
    \begin{subfigure}[b]{\linewidth}
        \centering
        \includegraphics[width=\linewidth]{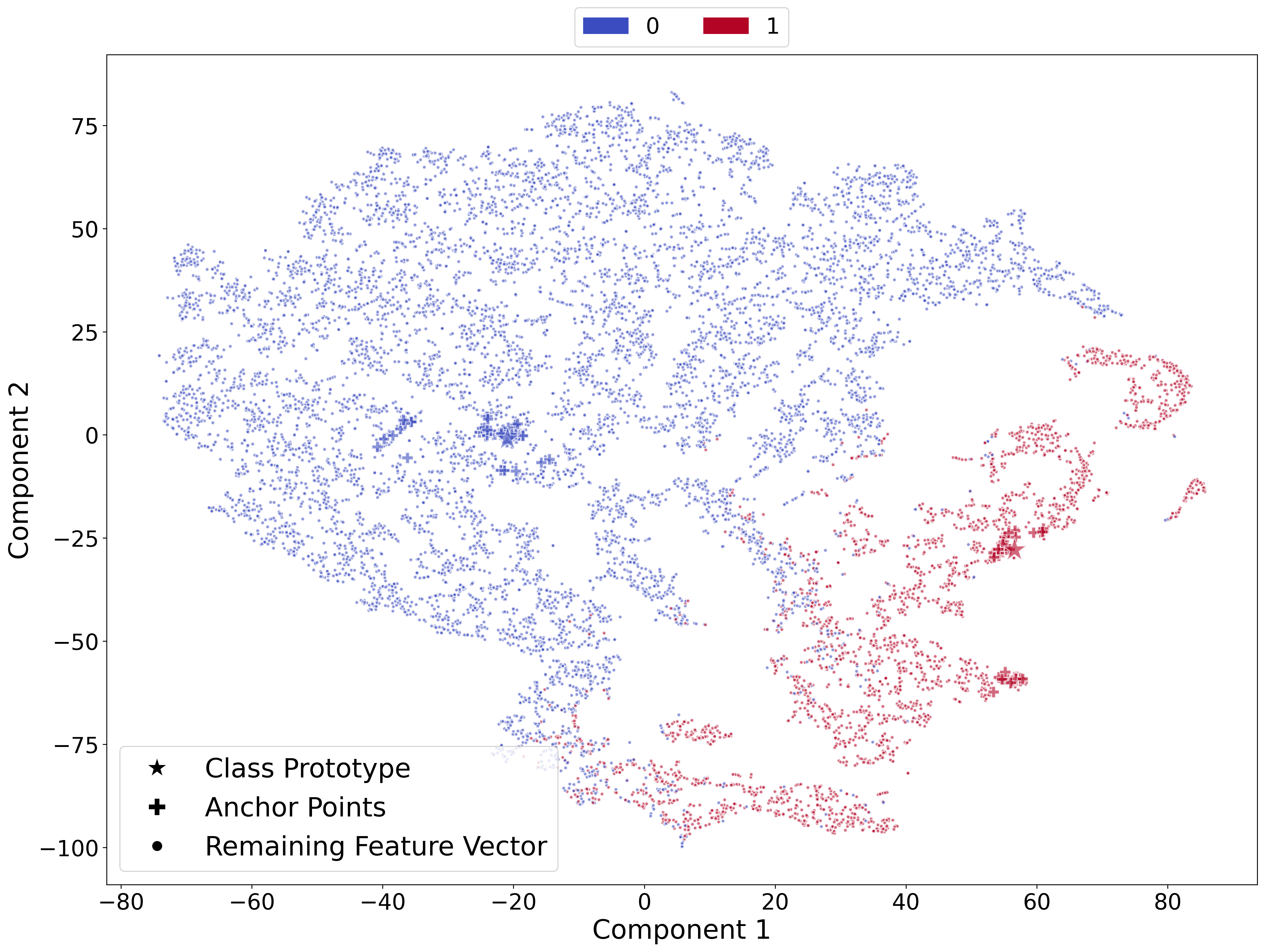}
        \caption{Anchor points selected through ``Closest" to prototype}
        \label{fig:randomnhie}
    \end{subfigure}
    \caption{\textbf{Visualization of Base Class Feature Space Distrbution for the NHIE Dataset}}
    \label{fig:nhietsne}
\end{figure}

\begin{figure}
    \centering
    \begin{subfigure}[b]{\linewidth}
        \centering
        \includegraphics[width=\linewidth]{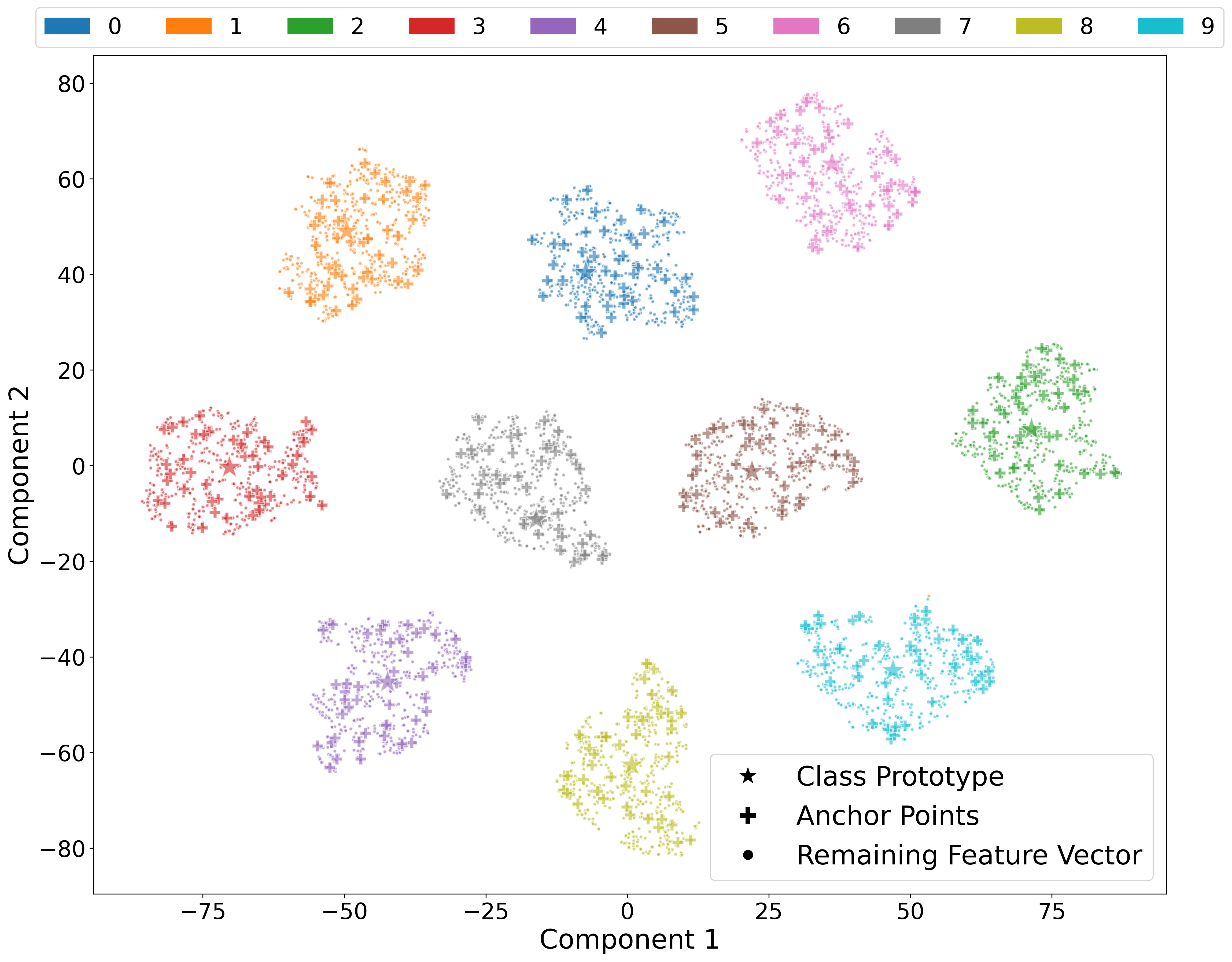}
        \caption{Anchor points selected as ``Random Sample"}
        \label{fig:closestgrabmyo}
    \end{subfigure}
    \hfill
    \begin{subfigure}[b]{\linewidth}
        \centering
        \includegraphics[width=\linewidth]{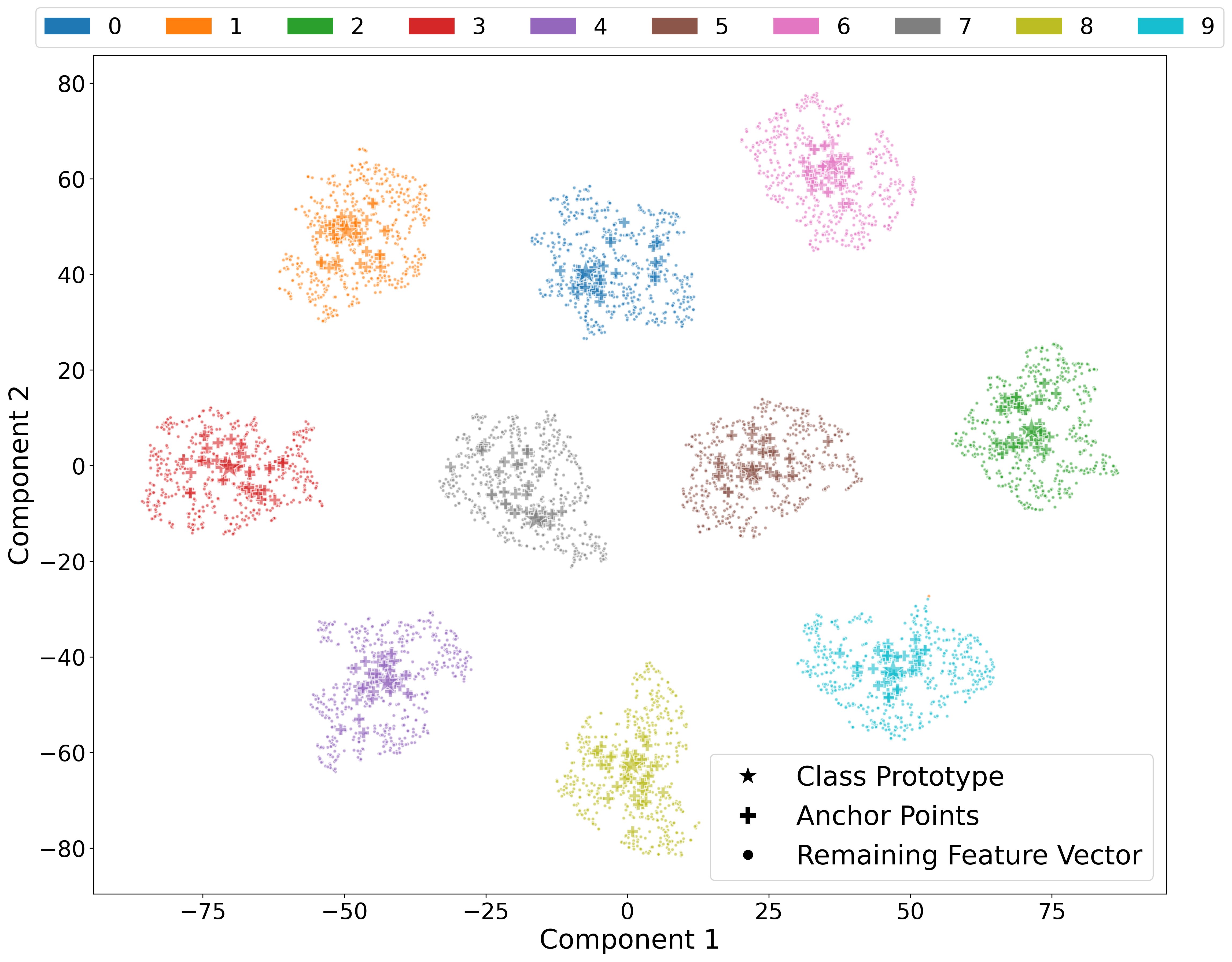}
        \caption{Anchor points selected through ``Closest" to prototype}
        \label{fig:randomgrabmyo}
    \end{subfigure}
    \caption{\textbf{Visualization of Base Class Feature Space Distrbution for GRABMyo Dataset}}
    \label{fig:grabmyotsne}
\end{figure}

We use t-SNE \cite{van2008visualizing} to visualize the distribution of the feature space of the training set and the selection of anchor points in the base session. We adopt the scikit-learn implementation with a perplexity of 30 and a max iteration of 1000 across all visualizations in Figure \ref{fig:bcitsne} and \ref{fig:nhietsne}
, \ref{fig:grabmyotsne}. The colors represent different class labels.

\section{Code Availability}
The code will be made publicly available upon publication.

\end{document}